\documentclass{article}

\usepackage[preprint]{neurips_2026}

\usepackage[utf8]{inputenc} 
\usepackage[T1]{fontenc}    
\usepackage{hyperref}       
\usepackage{url}            
\usepackage{booktabs}       
\usepackage{amsfonts}       
\usepackage{nicefrac}       
\usepackage{microtype}      
\usepackage{xcolor}         
\usepackage{graphicx}

\usepackage{multirow}
\usepackage{tabularx}
\usepackage{amsmath}
\usepackage{longtable}
\usepackage{subcaption}
\usepackage{changepage, threeparttable} 
\usepackage{url}
\usepackage{rotating}

\title{EpiCastBench: Datasets and Benchmarks for Multivariate Epidemic Forecasting}

\author{%
  Madhurima Panja\thanks{Equal Contribution} \\
  Sorbonne University Abu Dhabi\\
  \texttt{madhurima.panja@sorbonne.ae} \\
  \And
  Danny D'Agostino* \\
  Duke-NUS Medical School, Singapore\\
  \texttt{DannyDagostino@hotmail.com} \\
  \And
  Huitao Li \\
  Duke-NUS Medical School, Singapore\\
  \texttt{li.huitao@duke-nus.edu.sg} \\
  \And
  Tanujit Chakraborty\thanks{Corresponding Author} \\
  Sorbonne University Abu Dhabi\\
  \texttt{tanujit.chakraborty@sorbonne.ae} \\
  \And
  Nan Liu \\
   Duke-NUS Medical School, Singapore\\
  \texttt{liu.nan@duke-nus.edu.sg} \\
}

\begin{document}

\maketitle

\begin{abstract}
The increasing adoption of data-driven decision-making in public health has established epidemic forecasting as a critical area of research. Recent advances in multivariate forecasting models better capture complex temporal dependencies than conventional univariate approaches, which model individual series independently. Despite this potential, the development of robust epidemic forecasting methods is constrained by the lack of high-quality benchmarks comprising diverse multivariate datasets across infectious diseases and geographical regions. To address this gap, we present EpiCastBench, a large-scale benchmarking framework featuring 40 curated (correlated) multivariate epidemic datasets. These publicly available datasets span a wide range of infectious diseases and exhibit diverse characteristics in terms of temporal granularity, series length, and sparsity. We analyze these datasets to identify their global features and structural patterns. To ensure reproducibility and fair comparison, we establish standardized evaluation settings, including a unified forecasting horizon, consistent preprocessing pipelines, diverse performance metrics, and statistical significance testing. By leveraging this framework, we conduct a comprehensive evaluation of 15 multivariate forecasting models spanning statistical baselines to state-of-the-art deep learning and foundation models. All datasets and code are publicly available on Kaggle (\url{https://www.kaggle.com/datasets/aimltsf/epicastbench}) and GitHub (\url{https://github.com/aimltsf/EpiCastBench}).
\end{abstract}

\section{Introduction}

Epidemic forecasting has emerged as a cornerstone of epidemiology and health data science, playing a central role in supporting public health policy design, enabling timely interventions, and guiding efficient resource allocation \cite{rosenfeld2021epidemic}. By accurately predicting the progression of infectious diseases, it provides actionable insights that help mitigate the socioeconomic impact of rapidly evolving outbreaks. The importance of such forecasting capabilities has been repeatedly underscored by major public health crises, including the 1918 Spanish flu and more recent outbreaks such as COVID-19. These events collectively highlight the urgent need for reliable forecasting tools to support decision-making under uncertainty \cite{jones2008global}. Traditionally, mechanistic epidemiological models have been widely used for this purpose; however, in recent years, the field has undergone a paradigm shift towards data-driven approaches. These methods leverage historical surveillance data to capture complex and evolving outbreak patterns without relying on strong compartmental assumptions \cite{rodriguez2024machine}. As a result, data-driven forecasting models, including statistical learning, deep neural networks, and more recently, foundation models, have gained significant attention as flexible and scalable alternatives across a wide range of epidemic prediction tasks \cite{brooks2018nonmechanistic}.

Despite this progress, a significant challenge in epidemic forecasting research is the limited availability of comprehensive multivariate datasets. Unlike fields such as computer vision and natural language processing, which benefit from large-scale standardized repositories like ImageNet \cite{deng2009imagenet} and GLUE \cite{wang2018glue}, the epidemic forecasting domain lacks comparable benchmark datasets. While archives such as the Monash Time Series Forecasting Repository \cite{godahewa2021monash} provide unified collections for general-purpose time series data, similar large-scale and standardized efforts for epidemic forecasting remain scarce. As a result, most existing datasets are either univariate \cite{panja2023epicasting}, disease-specific \cite{barman2025epidemic}, or focused on specific geographical regions \cite{colon2023projecting}, limiting their ability to capture the diversity of transmission dynamics across diseases and environments. This lack of diversity restricts the development and evaluation of forecasting models that can generalize across multiple epidemic settings. Furthermore, it hinders meaningful comparisons across studies, as many approaches rely on region-specific datasets that are often constrained by privacy and accessibility concerns. Consequently, there is a need for a unified benchmark that captures diverse epidemic dynamics and enables consistent, reproducible evaluation of forecasting models.

\begin{figure}
    \centering
    \includegraphics[width=0.8\linewidth]{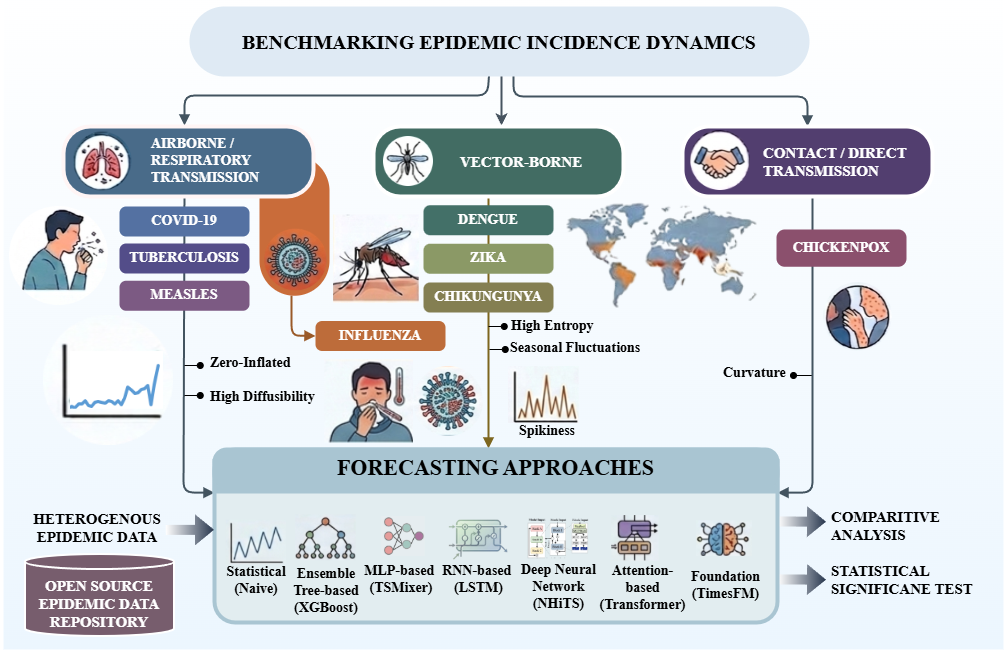}
    \caption{Overview of the EpiCastBench framework.}
    \label{fig_introduction_epidemics}
\end{figure}

This paper addresses this gap by introducing \textit{EpiCastBench}, visualized in Figure \ref{fig_introduction_epidemics}, as a comprehensive benchmark for evaluating data-driven epidemic forecasting models across diverse disease dynamics. EpiCastBench comprises 40 publicly available multivariate time series datasets of incidence cases, spanning multiple infectious diseases (COVID-19, chickenpox, chikungunya, dengue, influenza, measles, Zika, and tuberculosis), geographic regions, and transmission patterns. We implement standardized preprocessing pipelines and evaluate 15 representative forecasting models, including statistical methods such as Naive; machine learning models such as DLinear \cite{zeng2023transformers}, Random Forest \cite{breiman2001random}, XGBoost \cite{chen2016xgboost}, TSMixer \cite{chen2023tsmixer}, and KAN \cite{liu2025kan}; deep learning models including LSTM \cite{hochreiter1997long}, DeepAR \cite{salinas2020deepar}, TCN \cite{chen2020probabilistic}, N-BEATS \cite{oreshkin2020nbeats}, N-HiTS \cite{challu2023nhits}, Transformers \cite{vaswani2017attention}, and TiDE \cite{das2023longterm}; and foundation models such as Chronos-2 \cite{ansari2025chronos} and TimesFM \cite{das2024decoder}, all evaluated under a unified experimental setup across three forecasting horizons. We further adopt four widely used evaluation metrics and incorporate statistical significance testing to enable robust model comparisons. By integrating diverse datasets, modeling approaches, and evaluation setup into a single framework, EpiCastBench enables consistent model comparison and supports the development of more generalizable epidemic forecasting methods.

\section{Related work}

Data-driven approaches have shown significant promise in epidemic forecasting. Early methods primarily relied on mechanistic models of disease transmission, but their dependence on strong assumptions often limits generalization across real-world settings \cite{rodriguez2024machine}. To address these limitations, machine learning approaches such as ensemble-based Extreme Gradient Boosting (XGBoost) have been explored for their ability to capture nonlinear relationships and perform well on tabular epidemic data with limited samples \cite{lv2021time}. With the increasing availability of high-frequency epidemic datasets, particularly during the COVID-19 pandemic, advanced neural architectures such as Long Short-Term Memory (LSTM) networks \cite{chimmula2020time} and attention-based Transformers \cite{wu2020deep} have been widely adopted to model complex temporal dependencies and transmission patterns. More recently, foundation models have emerged as a promising direction, leveraging large-scale pretraining to enhance generalization across heterogeneous epidemic time series \cite{gong2025epillm}.

Recent works have focused on collecting and sharing epidemic datasets for advancing data-driven forecasting techniques \cite{adiga2026idobe, panja2023epicasting}. Notably, infectious disease forecasting challenges organized by the US CDC Epidemic Prediction Initiative have evaluated models across diseases such as influenza-like illness (ILI), dengue, and chikungunya \cite{biggerstaff2016results, johansson2019open}. Similar initiatives during the COVID-19 pandemic further highlighted the role of forecasting in public health decision-making \cite{cramer2022evaluation}. However, most existing studies focus on a specific disease and univariate formulations, limiting generalization across datasets and epidemiological contexts. This study builds on prior epidemic forecasting efforts while adopting the design principles of large-scale time series benchmarks \cite{godahewa2021monash}. We introduce EpiCastBench, a curated collection of multivariate epidemic datasets with standardized evaluation protocols and unified benchmarking. 
It supports reproducible evaluation and is designed to be extensible to future forecasting methods.

\section{Epidemic dataset} \label{section_epidemic_dataset}
This section details the diverse collection of epidemic datasets used in our large-scale benchmarking study. The EpiCastBench consists of 40 multivariate epidemic incidence datasets covering a wide spectrum of infectious diseases, including COVID-19, chickenpox, chikungunya, dengue, influenza, measles, tuberculosis, and Zika. These datasets are sourced from 27 distinct geographical regions worldwide, spanning countries at different stages of economic development, thereby capturing substantial spatial variability and epidemiological diversity. A key challenge in epidemic forecasting is the limited availability of long-term time series data. To address this, the EpiCastBench framework incorporates datasets with varying temporal lengths ranging from 60 to 2088 observations, thereby covering small, medium, and relatively large (in epidemiology context) sample-sized datasets. This variability enables the evaluation of forecasting models under realistic data availability conditions. Additionally, the dimensionality of the datasets varies substantially, with the number of epidemic incidence series ranging from 4 to 107. This benchmark also includes high-dimensional, low-sample-size scenarios, which are particularly relevant in epidemic modeling where data collection is often constrained. Specifically, the tuberculosis dataset of China comprises incidence observations from up to 31 provinces with only 60 time steps, while the chikungunya dataset of Brazil contains 72 temporal observations collected from 26 states. These epidemic datasets enable a comprehensive assessment of forecasting models across diverse data regimes, including settings with sparse observations and high variability. Furthermore, due to the intrinsic dynamics of infectious disease spread, several datasets exhibit prolonged periods of low or no reported cases between epidemic waves. This results in zero-inflated time series, where a significant proportion of observations are zeros across different regions. Such characteristics introduce additional challenges for forecasting models, particularly in accurately capturing both the occurrence and intensity of outbreaks. The datasets span multiple temporal granularities, including daily, weekly, and monthly frequencies. This heterogeneity reflects real-world epidemiological surveillance systems and allows us to evaluate the adaptability of forecasting models to varying sampling frequencies and temporal resolutions. In total, the benchmark includes 21 datasets on COVID-19, 9 on dengue, 2 on Zika virus, 2 on influenza, 2 on chikungunya, 2 on tuberculosis, and one dataset each for measles and chickenpox. All datasets are curated from official data repositories and previously published studies on epidemic modeling and forecasting, ensuring both reliability and reproducibility. A summary of these epidemic datasets, including the number of temporal observations, frequency, time span, feature count, proportion of zero observations, and their respective sources, is provided in Table \ref{tab:datasets_overview}. The table also reports the presence of missing observations, with six out of the 40 datasets containing missing values. In our study, missing values are imputed using the last observation carried forward (LOCF) approach, following \cite{godahewa2021monash}.

\begin{table}[h]
\caption{Overview of epidemic datasets considered in the benchmark study. In some instances, multiple datasets exist for the same location / country and disease, differing only in their temporal coverage. To distinguish between such cases, a numerical suffix (e.g., 2) is appended to the location name.}
\label{tab:datasets_overview}
\scriptsize 
\resizebox{\columnwidth}{!}{\begin{tabular}{cccccccccccccc} 
\toprule
	&	Disease	&	Location	&	Start date	&	End date	&	Frequency	&	Temporal Observations	&	Series count	&	Missing value count	& Zero-inflation(\%) &	Source	\\
    \midrule
1	&	\multirow{21}{*}{COVID-19}	&	Australia	&	\texttt{01/03/2020}	&	\texttt{19/10/2023}	&	Daily	&	1271	&	17	&	57	& 61.29 &	\cite{nsw2024}	\\
2	&		&	Belgium	&	\texttt{01/03/2020}	&	\texttt{03/10/2023}	&	Daily	&	1312	&	11	&	0 & 23.43	&	\cite{sciensanoCovid}	\\
3	&		&	Brazil	&	\texttt{28/03/2020}	&	\texttt{23/05/2021}	&	Daily	&	422	&	27	&	0 & 1.04	&	\cite{coronavirusBrazil}	\\
4	&		&	Canada	&	\texttt{23/01/2020}	&	\texttt{07/12/2020}	&	Daily	&	320	&	16	&	0 & 20.07	&	\cite{dong2020}	\\
5	&		&	Canada 2	&	\texttt{08/02/2020}	&	\texttt{25/05/2024}	&	Weekly	&	225	&	13	&	0	& 62.85 &	\cite{canadaCovid19}	\\
6	&		&	Chile	&	\texttt{16/03/2020}	&	\texttt{20/06/2020}	&	Daily	&	97	&	15	&	0	& 71.48 & \cite{silvaAllendeCovid19}	\\
7	&		&	China	&	\texttt{23/01/2020}	&	\texttt{07/12/2020}	&	Daily	&	320	&	33	&	0	& 75.25 &	\cite{dong2020}	\\
9	&		&	Colombia	&	\texttt{03/09/2020}	&	\texttt{07/14/2020}	&	Daily	&	127	&	33	&	1	& 45.22 &	\cite{colombiaCovid19}	\\
8	&		&	Czech	&	\texttt{01/03/2020}	&	\texttt{31/05/2024}	&	Daily	&	1542	&	14	&	11	& 11.23 &	\cite{czechCovid19}	\\
10	&		&	EU	&	\texttt{01/01/2020}	&	\texttt{26/10/2022}	&	Daily	&	1030	&	30	&	0	& 11.45 &	\cite{ecdcCovid19}	\\
11	&		&	Germany	&	\texttt{03/03/2020}	&	\texttt{26/01/2023}	&	Daily	&	1060	&	16	&	0 & 1.35 &	\cite{germanyCovid19}	\\
12	&		&	India	&	\texttt{26/04/2020}	&	\texttt{31/10/2021}	&	Daily	&	554	&	31	&	0	&	0.09 & \cite{indiaCovid19}	\\
13	&		&	Ireland	&	\texttt{01/03/2020}	&	\texttt{13/11/2023}	&	Daily	&	1353	&	26	&	0	& 21.09 &	\cite{irelandCovid19}	\\
14	&		&	Italy	&	\texttt{25/02/2020}	&	\texttt{15/05/2024}	&	Daily	&	1542	&	107	&	0	& 11.85 &	\cite{italyCovid19}	\\
15	&		&	Japan	&	\texttt{16/01/2020}	&	\texttt{08/05/2023}	&	Daily	&	1209	&	47	&	0	&	19.45 & \cite{japanCovid19}	\\
16	&		&	Mexico	&	\texttt{26/02/2020}	&	\texttt{24/06/2023}	&	Daily	&	1215	&	32	&	0	& 2.58	& \cite{mexicoCovid19}	\\
17	&		&	Netherlands	&	\texttt{04/10/2021}	&	\texttt{31/03/2023}	&	Daily	&	544	&	12	&	0	& 0.43 &	\cite{netherlandsCovid19}	\\
18	&		&	Spain	&	\texttt{21/03/2020}	&	\texttt{24/05/2020}	&	Daily	&	65	&	19	&	0	&	7.13 & \cite{spainCovid19}	\\
19	&	&	Switzerland	&	\texttt{02/03/2020}	&	\texttt{27/05/2024}	&	Weekly	&	222	&	29	&	0	& 5.17 &	\cite{switzerlandCovid19}	\\
20	&		&	UK	&	\texttt{03/02/2020}	&	\texttt{01/08/2020}	&	Daily	&	181	&	4	&	0	& 27.07 &	\cite{ukCovid19}	\\
21	&		&	US	&	\texttt{22/01/2020}	&	\texttt{23/03/2023}	&	Daily	&	1157	&	56	&	0	& 30.28 &	\cite{usCovid19}	\\ \hline
22	&	Chickenpox	&	Hungary	&	\texttt{03/01/2005}	&	\texttt{29/12/2014}	&	Weekly	&	522	&	20	&	0	& 4.32 &	\cite{rozemberczki2021chickenpox}	\\ \hline
23	&	\multirow{2}{*}{Chikungunya}	&	Brazil	&	\texttt{01/01/2015}	&	\texttt{01/01/2021}	&	Monthly	&	72	&	27	&	1	& 45.31 &	\cite{daSilva2022}	\\
24	&		&	Colombia	&	\texttt{09/06/2014}	&	\texttt{04/07/2016}	&	Weekly	&	101	&	28	&	8	& 2.98 &	\cite{charniga2021}	\\ \hline
25	&	\multirow{9}{*}{Dengue}	&	Argentina	&	\texttt{22/09/2019}	&	\texttt{20/06/2021}	&	Weekly	&	92	&	23	&	0	& 19.85 &	\cite{clarke2024}	\\
26	&		&	Brazil	&	\texttt{01/01/2000}	&	\texttt{01/12/2019}	&	Monthly	&	240	&	27	&	0 & 5.58	&	\cite{lowe2021}	\\
27	&		&	Colombia	&	\texttt{31/12/2006}	&	\texttt{25/12/2022}	&	Weekly	&	835	&	33	&	0	& 0.48 & \cite{clarke2024}	\\
28	&		&	Malaysia	&	\texttt{22/02/2010}	&	\texttt{28/12/2015}	&	Weekly	&	305	&	15	&	1	&	16.78 & \cite{rDengueAnalysis}	\\
29	&		&	Malaysia 2	&	\texttt{29/12/2019}	&	\texttt{07/08/2022}	&	Weekly	&	137	&	15	&	0	& 10.41 &	\cite{pahangWDF}	\\
30	&		&	Panama	&	\texttt{31/12/2017}	&	\texttt{06/11/2022}	&	Weekly	&	254	&	11	&	0	&	37.25 & \cite{clarke2024}	\\
31	&		&	Peru	&	\texttt{31/07/2011}	&	\texttt{22/12/2013}	&	Weekly	&	126	&	19	&	0	&	31.07 & \cite{clarke2024}	\\
32	&		&	Philippines	&	\texttt{01/01/1994}	&	\texttt{01/12/2010}	&	Monthly	&	204	&	65	&	0	& 21.67 &	\cite{clarke2024}	\\
33	&		&	Taiwan	&	\texttt{11/05/2014}	&	\texttt{13/03/2016}	&	Weekly	&	97	&	21	&	0	&	23.07 &\cite{clarke2024}	\\ \hline
34	&	\multirow{2}{*}{Influenza}	&	US	&	\texttt{03/10/2010}	&	\texttt{20/09/2020}	&	Weekly	&	521	&	53	&	0	& 4.20 & 	\cite{morris2023}	\\
35	&		&	US 2	&	\texttt{28/09/1997}	&	\texttt{20/09/2020}	&	Weekly	&	1200	&	10	&	0	& 8.35 &	\cite{morris2023}	\\ \hline
36	&	Measles	&	US	&	\texttt{01/01/1940}	&	\texttt{31/12/1979}	&	Weekly	&	2088	&	50	&	0	& 1.67 & 	\cite{projectTycho}	\\ \hline
37	&	\multirow{2}{*}{Zika}	&	Colombia	&	\texttt{10/08/2015}	&	\texttt{12/06/2017 }	&	Weekly	&	97	&	29	&	0	&	3.27 & \cite{charniga2021}	\\
38	&		&	Mexico	&	\texttt{28/11/2015}	&	\texttt{30/06/2018}	&	Weekly	&	136	&	29	&	0	&	74.69 & \cite{cdcepiZika}	\\ \hline
39	&	\multirow{2}{*}{Tuberculosis}	&	Japan	&	\texttt{01/01/1998}	&	\texttt{01/12/2015}	&	Monthly	&	216	&	47	&	0	& 0.00 &	\cite{barman2025epidemic}	\\
40	&		&	China	&	\texttt{01/01/2014}	&	\texttt{31/12/2018}	&	Monthly	&	60	&	31	&	0	& 0.00 &	\cite{barman2025epidemic}	\\

\bottomrule
\end{tabular}}
\end{table}

\subsection{Global feature analysis} \label{section_feature_analysis}
\begin{figure}
    \centering
    \includegraphics[width=0.7\linewidth]{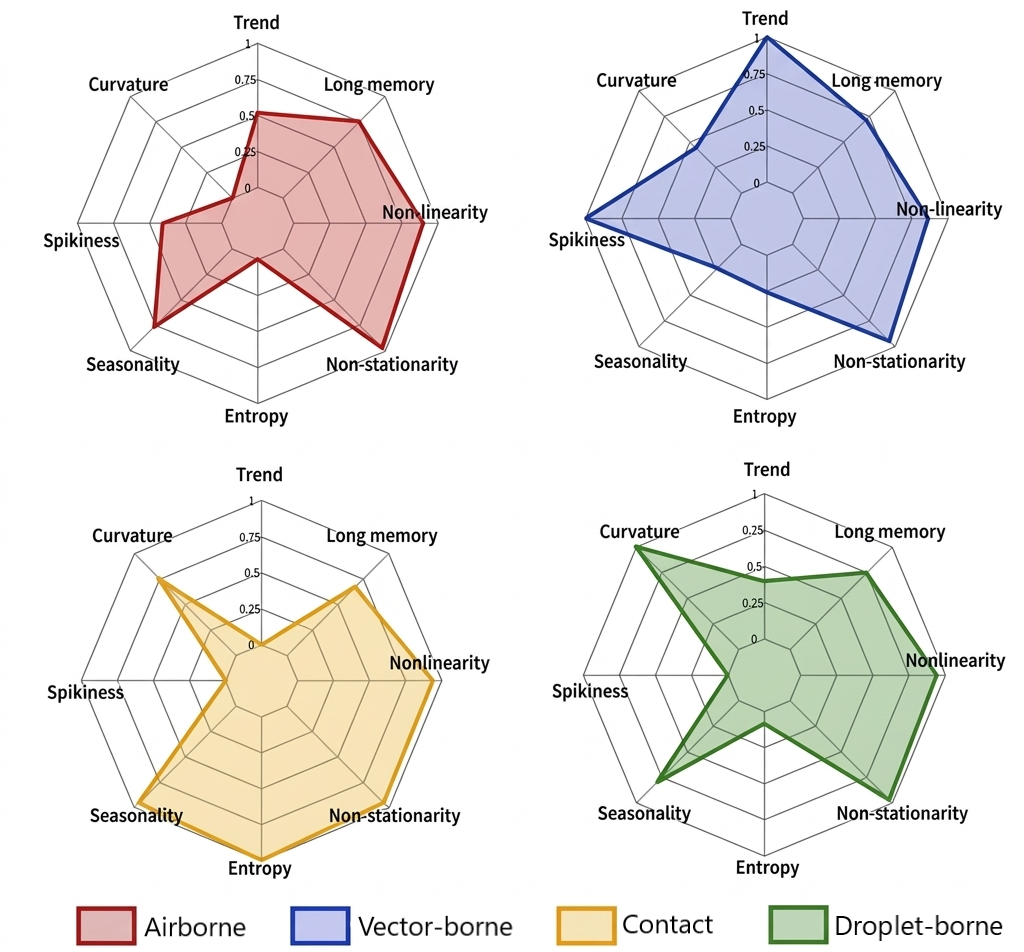}
    \caption{Radar plot comparing global features of epidemic datasets across transmission channels.}
    \label{fig_global_patterns}
\end{figure}

We analyze the epidemic time series datasets considered in this study to better understand their overall behavior and to characterize them based on underlying transmission patterns. This analysis enables the selection of forecasting models that are better suited to the dynamics of different disease categories. Specifically, we examine eight global time series features, including trend, long memory, nonlinearity, non-stationarity, entropy, seasonality, spikiness, and curvature, using the \textit{tsfeatures} package in \textbf{R} \cite{hyndman2023tsfeatures}. Figure \ref{fig_global_patterns} presents a comprehensive visualization of these normalized characteristics across datasets grouped by transmission type. The results reveal clear distinctions among the different categories. Airborne diseases, such as COVID-19, measles, and tuberculosis, exhibit strong nonlinearity and non-stationarity, indicating complex and evolving transmission dynamics that are not adequately captured by linear models. These datasets also show moderate levels of long memory, trend, spikiness, and seasonality, suggesting that future observations depend on past values and that some periodic behavior is present. Additionally, these epidemic time series exhibit lower entropy and curvature, suggesting that despite their inherent complexity, they do not display highly irregular behavior. In contrast, vector-borne diseases, including dengue, chikungunya, and Zika, are characterized by pronounced spikiness and a strong trend component. This reflects the presence of sudden outbreaks, often influenced by environmental and ecological factors such as vector population dynamics. The relatively high long memory suggests persistence in the system, while the combination of nonlinearity and non-stationarity points to a balance between structured patterns and dynamic variability. The comparatively lower seasonality may indicate that, although vector activity is inherently seasonal, aggregation effects tend to smooth out strict periodic cycles in the observed data. For the contact-based chickenpox time series, the high entropy and strong seasonality indicate a combination of randomness and periodic structure. The significantly high nonlinearity and non-stationarity further suggest complex transmission mechanisms, likely driven by human interactions and behavioral patterns. In addition, lower spikiness implies that outbreaks evolve more gradually, while higher curvature reflects more noticeable changes over the temporal trajectory. Finally, droplet-borne diseases such as influenza display relatively balanced characteristics, with moderately high nonlinearity, non-stationarity, and seasonality. A distinguishing feature in this category is higher curvature, indicating more rapid changes in the temporal pattern, likely due to intervention effects or fast transmission cycles. Compared to vector-borne diseases, spikiness is lower, suggesting fewer abrupt peaks, while moderate entropy reflects a mix of structured and stochastic behavior. Overall, this analysis highlights that different transmission mechanisms are associated with distinct temporal behaviors. These differences in the global characteristics pose critical challenges for forecasting, emphasizing that the model selection should be guided by the intrinsic dynamics of each disease category.

\section{Baseline evaluation}
This section details the empirical setup used for benchmarking the baseline models and describes the performance of the forecasting frameworks.

\subsection{Baseline evaluation setup}
In epidemic forecasting, it is important to benchmark models spanning from simple baselines to advanced foundation models, as simple approaches can remain competitive and interpretable while more complex models capture intricate epidemic dynamics \cite{godahewa2021monash}. However, existing studies often lack standardized and reproducible evaluation settings designed for epidemic data. To address this, we benchmark 15 multivariate forecasting models ranging from classical baselines to state-of-the-art approaches under a fixed-origin evaluation protocol across all datasets. This setup enables direct and fair comparison of newly proposed methods against a consistent set of reference models.

The evaluated models include statistical baselines such as the Naive model; machine learning approaches including decomposition-based linear (DLinear) \cite{zeng2023transformers}, Random Forest \cite{breiman2001random}, and XGBoost \cite{chen2016xgboost}; neural architectures such as Time Series Mixer (TSMixer) \cite{chen2023tsmixer} and Kolmogorov Arnold Networks (KAN) \cite{liu2025kan}; deep learning models including recurrent architectures such as LSTM \cite{hochreiter1997long} and Deep Autoregressive models (DeepAR) \cite{salinas2020deepar}, convolution-based Temporal Convolutional Networks (TCN) \cite{chen2020probabilistic}, feed-forward architectures such as Neural Basis Expansion Analysis for Interpretable Time Series Forecasting (NBeats) \cite{oreshkin2020nbeats} and Neural Hierarchical Interpolation for Time Series Forecasting (NHiTS) \cite{challu2023nhits}, encoder-decoder-based models such as Transformers \cite{vaswani2017attention} and time series Dense Encoder (TiDE) \cite{das2023longterm}; and foundation models including Chronos-2 \cite{ansari2025chronos} and TimesFM \cite{das2024decoder}. These multivariate forecasting models collectively comprise a diverse set of approaches across statistical, machine learning, deep learning, and foundation model paradigms. A detailed description of each baseline model is provided in Appendix. In this benchmarking, we exclude classical multivariate models, such as Vector Autoregression (VAR), because the epidemic time series exhibit zero-inflated behavior. The prolonged periods of zero or near-zero case counts in these series often lead to numerical instability and unreliable parameter estimation in the VAR model, making them unsuitable for consistent evaluation across all datasets. 

To evaluate model performance across different forecasting horizons, we consider three settings: long-term, medium-term, and short-term forecasting. Long-term forecasting captures the overall trajectory of the epidemic and supports strategic planning, while medium-term forecasting focuses on intermediate dynamics such as emerging waves. On the contrary, short-term forecasting evaluates the ability to predict immediate fluctuations, which is critical for operational decision-making. We adopt a rolling-window evaluation scheme, where forecasting horizons are defined as 30, 14, and 7 days for daily data; 12, 8, and 4 weeks for weekly data; and 24, 12, and 6 months for monthly data. Additionally, model performance is assessed using scale-independent measures, such as Mean Absolute Scaled Error (MASE) and symmetric Mean Absolute Percentage Error (SMAPE), as well as scale-dependent metrics including Mean Absolute Error (MAE) and Root Mean Squared Error (RMSE) \cite{godahewa2021monash}. The formal definitions of these metrics are provided in Appendix. By definition, lower values of these metrics indicate better forecasting performance. 

In EpiCastBench, the implementation of all forecasting models and their performance evaluation are conducted in \textbf{Python} using the \textit{Darts} library \cite{herzen2022darts}, which provides a unified, reproducible framework. Each model is trained using a historical input window of 24 observations to predict the next 12 observations in a forecasting chunk, with 50 epochs of training and a fixed random state of 42. All other parameters are kept at their default settings to establish these models as reference benchmarks, while acknowledging that additional hyperparameter tuning and task-specific adjustments could further improve their performance. In the evaluation, long-term forecasting is conducted for all datasets except the daily COVID-19 cases of Spain, which contain only 65 observations per series, making the 30-day-ahead evaluation unreliable. Medium and short-term evaluations are performed across all 40 datasets. Additionally, SMAPE is not reported for the Malaysia dengue dataset, and MASE is not reported for medium and long-term forecasting of COVID-19 cases in Chile and Canada due to high zero-inflation in the test period, which leads to unstable metric values.

\subsection{Baseline performance}
\begin{figure}
    \centering
    \includegraphics[width=0.9\linewidth]{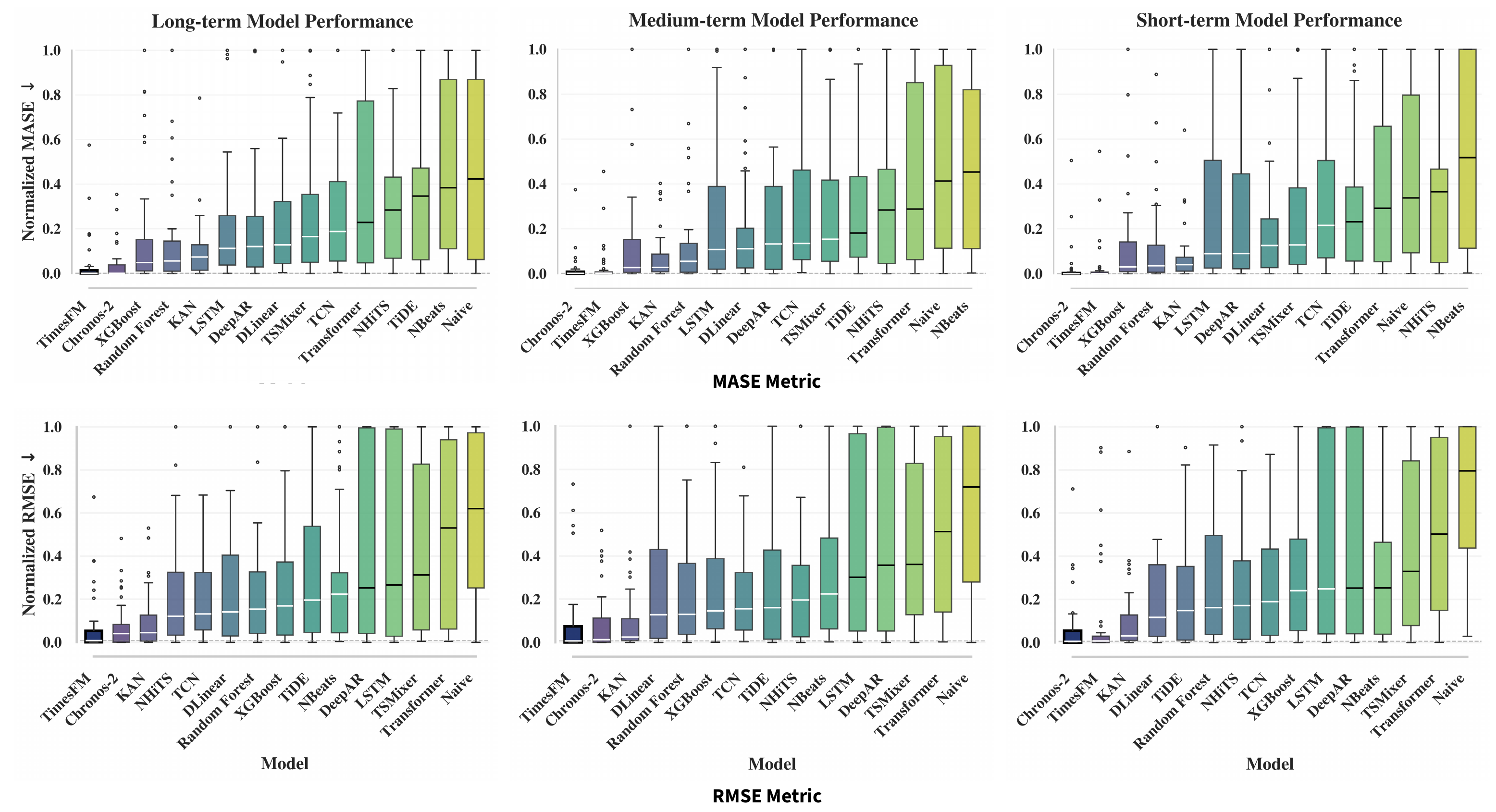}
    \caption{Boxplots comparing model performance using MASE (top) and RMSE (bottom) across long (left), medium (middle), and short-term (right) forecast horizons.}
    \label{Fig_Bixplot_MASE_RMSE}
\end{figure}

\begin{figure}
    \centering
    \includegraphics[width=1\linewidth]{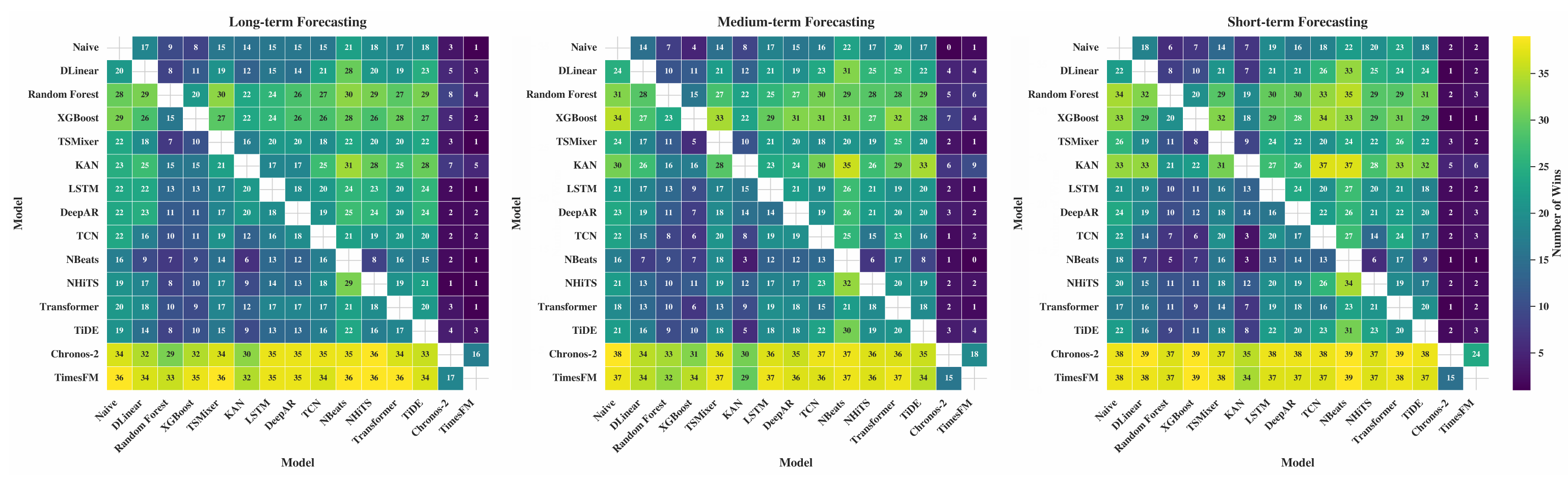}
    \caption{Win-count heatmaps based on the MASE metric for long (left), medium (middle), and short-term (right) forecasting horizons. Each cell $(i, j)$ represents the number of datasets where the model in row $i$ outperformed the model in column $j$. Brighter colors (yellow) represent stronger pairwise dominance, while darker colors (purple) indicate a lower frequency of wins.}
    \label{fig_win_heatmap}
\end{figure}
Figure \ref{Fig_Bixplot_MASE_RMSE} summarizes the performance of the baseline models by illustrating the distribution of their normalized MASE and RMSE metrics for long, medium, and short-term forecasting horizons. 
Across all forecasting tasks and evaluation metrics, the foundation models, particularly TimesFM and Chronos-2, demonstrate superior performance, consistently achieving the lowest median errors along with relatively smaller interquartile ranges. This indicates that foundation models provide both accurate forecasts and consistent behavior across diverse epidemic datasets. Their advantage is especially evident in long-term forecasting, where the ability to capture complex temporal dependencies, delayed effects, and non-stationary dynamics becomes critical. This behavior can be attributed to their pretraining on large-scale time series corpora, which enables better generalization to unseen epidemic dynamics. 

Classical machine learning models, including KAN, Random Forest, XGBoost, and DLinear, show competitive performance across all horizons and, in several cases, outperform deep learning models. These conventional architectures perform competitively with foundation models in short-term forecasting, where local temporal structures dominate and simpler relationships are sufficient to generate reliable forecasts. However, their performance gradually deteriorates as the forecasting horizon increases, reflecting their limited capacity to capture long-range dependencies and dynamic temporal patterns of epidemic time series. Among the deep learning models, LSTM, DeepAR, TCN, and Transformer exhibit moderate performance, but with noticeably higher variability, as indicated by wider interquartile ranges and the presence of outliers. This suggests a sensitivity to dataset characteristics and training dynamics, which is particularly relevant in epidemic forecasting scenarios where data can be sparse, noisy, and highly heterogeneous. 
Models such as NBeats and NHiTS show horizon and metric-dependent performance. For MASE, NBeats has higher median errors and greater variability, indicating lower stability, whereas NHiTS also has high median errors but comparatively smaller interquartile ranges and notable outliers. For RMSE, both models have similar performance for short and medium-term forecasting, with moderate errors and low variability. However, in long-term forecasting, NHiTS becomes more competitive, approaching foundation models, likely due to its ability to capture multi-scale temporal patterns, while NBeats remains relatively consistent across horizons. Additionally, the Naive baseline consistently produces the highest median errors across all forecasting horizons, accompanied by significant variability. This highlights the non-trivial nature of the epidemic forecasting task and further validates the effectiveness of more advanced modeling approaches. The forecasting performance observed for the SMAPE and MAE metrics (reported in Appendix) follows similar patterns to those of MASE and RMSE, with the foundation models consistently outperforming both baseline and state-of-the-art approaches in terms of lower median errors and reduced variability. Furthermore, we provide detailed performance results for each individual epidemic dataset, horizon, and performance metric in Appendix. These results show that disease type is a key factor in determining model performance, with foundation models excelling on airborne and droplet diseases with more regular temporal structure, while vector-borne and contact-based diseases with more irregular and variable dynamics are better captured by classical models. They further reveal that the impact of zero inflation is closely linked to the size of the dataset. In particular, while foundation models maintain strong and stable performance in moderate and relatively large datasets, even in the presence of substantial sparsity, their advantage becomes less consistent in small, highly zero-inflated time series. In such settings, conventional machine learning models can occasionally outperform more complex frameworks, particularly in short-term forecasting. As the sample size increases and longer forecasting horizons are considered, the robustness of foundation models to zero inflation becomes more evident, leading to more consistent performance across datasets. Overall, the empirical results indicate that while foundation models (TimesFM and Chronos-2) and classical machine learning frameworks such as KAN, XGBoost, Random Forest, and DLinear generally achieve strong performance, their relative advantages are dependent on the forecasting horizon and the underlying epidemic dynamics. 

\paragraph{Win-count Analysis:} To assess the relative performance of forecasting models across diverse epidemic datasets, we conduct a pairwise win-count analysis based on the MASE metric. The resulting heatmap, shown in Figure \ref{fig_win_heatmap}, demonstrates model dominance by recording the number of datasets where the $i^{th}$ model (row) achieves a lower MASE than the $j^{th}$ model (column). The observed patterns closely align with the overall performance evaluation, where the foundation models, TimesFM and Chronos-2, consistently achieve the most wins against both classical and deep learning baselines. This dominance is visually reflected by the prominent yellow bands corresponding to these models, indicating their strong and consistent superiority across datasets. In contrast, the concentration of darker regions in the upper portions of the heatmap highlights the performance gap between traditional baselines and advanced foundation architectures. 

\begin{figure}
    \centering
    \includegraphics[width=0.65\linewidth]{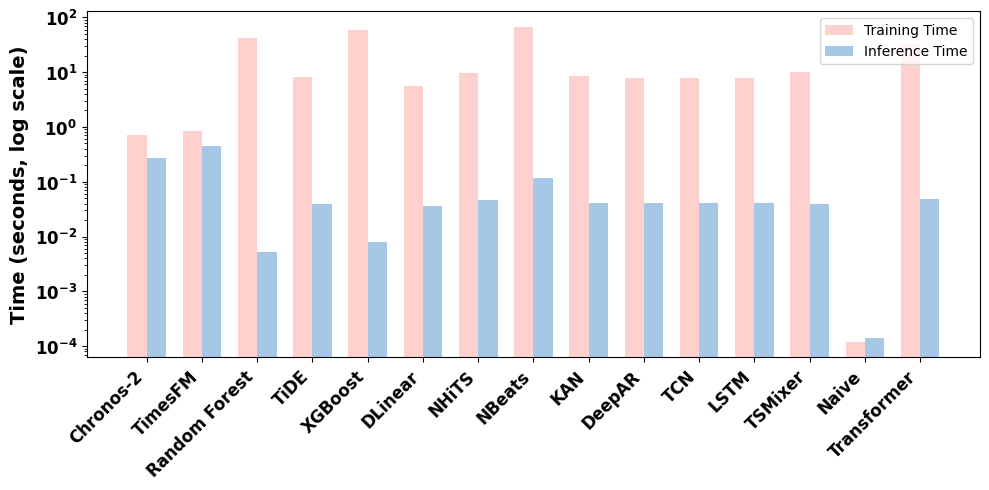}
    \caption{Training and inference times of all models for the short-term forecasting task on the Belgium COVID-19 dataset, run on an H100 High-RAM GPU using Google Colab. Models on the x-axis are ranked according to MASE performance, while the y-axis shows computation time (in seconds) on a logarithmic scale.}
    \label{Fig_Execution_time}
\end{figure}

\paragraph{Execution Time:} Figure \ref{Fig_Execution_time} reports the training and inference times for short-term prediction of Belgium COVID-19 cases, with models arranged on x-axis by MASE ranking. All experiments are conducted on an H100 High-RAM GPU in Google Colab, and execution times (in seconds) are reported on a logarithmic scale. The Naive model achieves the lowest computational cost but does not capture temporal structure, instead rely on persistence of the last observation. Among machine learning methods, tree-based models incur higher training costs due to ensemble fitting, while maintaining relatively efficient inference and competitive performance. In contrast, DLinear, KAN, and TSMixer exhibit lower training cost but increased inference time than ensemble frameworks, with limited improvements in forecasting performance. The deep learning models such as TiDE, NHiTS, DeepAR, TCN, and LSTM show stable computation, while more complex architectures like NBeats and Transformer require higher training time without consistent performance gains. On the other hand, foundation models exhibit a distinct computational profile. Due to their large-scale pretraining and ability to leverage contextual learning in a zero-shot manner, they require lower training time compared to both classical machine learning and deep learning models. However, inference time is higher due to autoregressive generation and contextual conditioning mechanisms. A similar execution time pattern is observed across datasets, with relative computational differences largely consistent across models and primarily driven by architecture and dataset size. Overall, the foundation models offer a favorable trade-off between training efficiency and predictive performance, with strong scalability for multivariate epidemic forecasting. 

\begin{figure}
    \centering
    \includegraphics[width=1.0\linewidth]{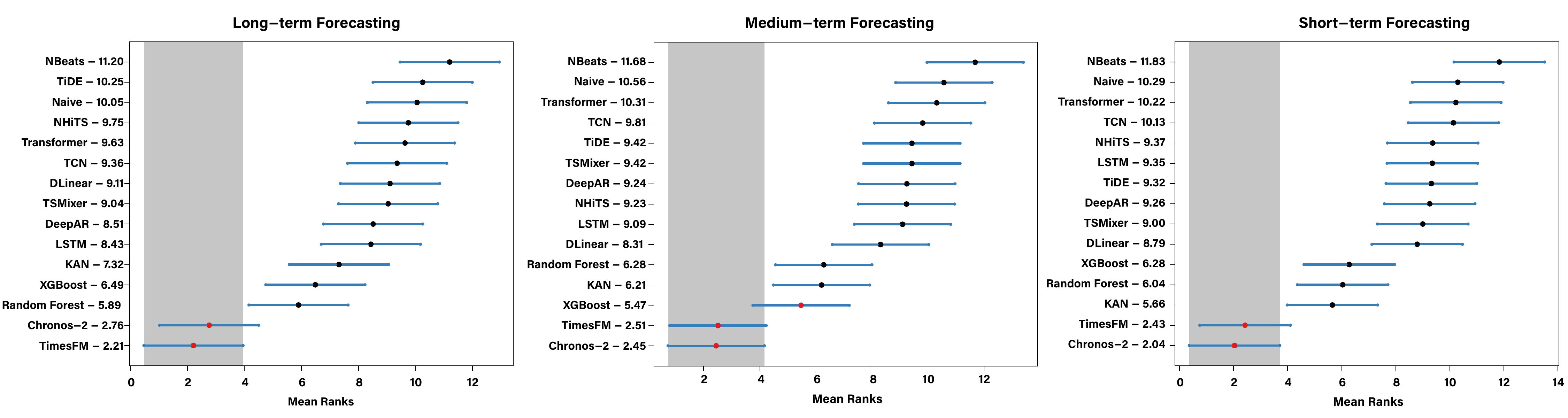}
    \caption{MCB test result for long (left), medium (middle), and short-term (right) forecasting horizons based on MASE. In the plot, `TimesFM - 2.21' indicates the average rank of the TimesFM framework for long-term forecasting is 2.21; a similar interpretation holds for other models and horizons.}
    \label{MCB_MASE_Plot}
\end{figure}

\subsection{Statistical significance test}

To assess the statistical significance of performance differences across the evaluated forecasting models, we employ the Friedman test \cite{friedman1937use, friedman1940comparison}. This non-parametric test is particularly well-suited for heterogeneous real-world epidemic time series where the normality and homoscedasticity assumptions are frequently violated. The Friedman test evaluates the null hypothesis that all models perform equivalently by computing their average ranks across multiple forecasting tasks. We conducted separate tests for each forecast horizon across all four performance indicators; in all instances, we obtained p-values significantly below 0.01. This consistent rejection of the null hypothesis confirms that the observed variations in model performance are statistically significant. Furthermore, to evaluate the robustness of the top-performing models, we conduct a post-hoc Multiple Comparisons with the Best (MCB) test \cite{koning2005m3}. This distribution-free procedure identifies models whose performance is statistically comparable with the `best' model based on the computed critical distance (CD). The MCB test results, visualized in Figure \ref{MCB_MASE_Plot} using the MASE metric, reveal that TimesFM emerges as the superior framework for long-term forecasting, while Chronos-2 achieves the lowest average ranks for medium and short-term horizons. The CD of these `best' performing models serves as the reference value of the test. Notably, across all the horizons, the CDs of the foundation models overlap within the reference threshold, indicating that differences among the foundation models are not statistically significant in the context of epidemic forecasting. Conversely, while baseline models such as XGBoost, KAN, and Random Forest achieve relatively lower ranks than other baselines, their CDs typically fall beyond the reference value of the `best' performing models. Other baseline architectures exhibit higher average ranks, demonstrating significant differences from the foundation models. These patterns remain consistent across additional error metrics (detailed in Appendix), underscoring the robustness of foundation models while highlighting that relative performances are sensitive to the specific forecasting horizon.

\section{Conclusion}
This paper presents EpiCastBench, a large-scale benchmark of 40 publicly available multivariate epidemic datasets for evaluating forecasting models, with a focus on reproducibility and scientific relevance. By leveraging standardized evaluation settings and widely adopted metrics, we compare 15 statistical, machine learning, deep learning, and foundation models across multiple datasets, metrics, and forecasting horizons. We further characterize the global features of epidemic time series based on their transmission dynamics. Notably, the study highlights challenges arising from complex temporal features and limited sample sizes, which are common in epidemic settings and significantly impact model performance. Overall, the EpiCastBench framework provides a comprehensive and reproducible evaluation setup, enabling systematic comparison and validation of new forecasting approaches on diverse multivariate epidemic datasets.

\paragraph{Limitations and Future Work:} While this study provides a standardized framework for evaluating multivariate epidemic forecasting models, several limitations remain. First, although EpiCastBench is designed for multivariate forecasting, it does not explicitly capture spatial dependencies and can be extended to spatiotemporal settings to model interactions in disease transmission across regions. Second, the current setup focuses on deterministic forecasting methods and evaluation metrics, without incorporating probabilistic forecasting or distribution-based evaluation strategies. As a promising future direction, we aim to address these limitations by extending the benchmark to a comprehensive spatiotemporal epidemic forecasting framework, which would complement the current version and enable more realistic modeling of disease propagation across space and time. In addition, incorporating probabilistic forecasting is an important direction, as it can better support informed decision-making under uncertainty in public health settings. Furthermore, while the current benchmark primarily evaluates data-driven forecasting models, recent studies have proposed epidemic-informed data-driven approaches that explicitly incorporate mechanistic epidemiological principles; integrating such models into the benchmark would provide a more comprehensive evaluation landscape.

\bibliographystyle{abbrv}
\bibliography{bibliography}

@article{jones2008global,
  title={Global trends in emerging infectious diseases},
  author={Jones, Kate E and Patel, Nikkita G and Levy, Marc A and Storeygard, Adam and Balk, Deborah and Gittleman, John L and Daszak, Peter},
  journal={Nature},
  volume={451},
  number={7181},
  pages={990--993},
  year={2008},
  publisher={Nature Publishing Group}
}

@inproceedings{deng2009imagenet,
  title={Imagenet: A large-scale hierarchical image database},
  author={Deng, Jia and Dong, Wei and Socher, Richard and Li, Li-Jia and Li, Kai and Fei-Fei, Li},
  booktitle={2009 IEEE conference on computer vision and pattern recognition},
  pages={248--255},
  year={2009},
  organization={Ieee}
}

@article{wang2018glue,
  title={Glue: A multi-task benchmark and analysis platform for natural language understanding},
  author={Wang, Alex},
  journal={arXiv preprint arXiv:1804.07461},
  year={2018}
}

@misc{rozemberczki2021chickenpox,
      title={{Chickenpox Cases in Hungary: a Benchmark Dataset for Spatiotemporal Signal Processing with Graph Neural Networks}}, 
      author={Benedek Rozemberczki and Paul Scherer and Oliver Kiss and Rik Sarkar and Tamas Ferenci},
      year={2021},
      eprint={2102.08100},
      archivePrefix={arXiv},
      primaryClass={cs.LG}
}

@article{daSilva2022,
  author    = {da Silva Neto, S.R. and Tabosa de Oliveira, T. and Teixiera, I.V. and others},
  title     = {Arboviral disease record data - Dengue and Chikungunya, Brazil, 2013–2020},
  journal   = {Scientific Data},
  volume    = {9},
  pages     = {198},
  year      = {2022},
  doi       = {10.1038/s41597-022-01252-8}
}

@article{charniga2021,
  author    = {Charniga, K. and Cucunubá, Z. M. and Mercado, M. and Prieto, F. and Ospina, M. and Nouvellet, P. and Donnelly, C. A.},
  title     = {Spatial and temporal invasion dynamics of the 2014--2017 Zika and chikungunya epidemics in Colombia},
  journal   = {PLOS Computational Biology},
  volume    = {17},
  number    = {7},
  pages     = {e1009174},
  year      = {2021},
  doi       = {10.1371/journal.pcbi.1009174}
}

@article{dong2020,
  author    = {Dong, E. and Du, H. and Gardner, L.},
  title     = {An interactive web-based dashboard to track COVID-19 in real time},
  journal   = {The Lancet. Infectious diseases},
  volume    = {20},
  number    = {5},
  pages     = {533--534},
  year      = {2020},
  doi       = {10.1016/S1473-3099(20)30120-1}
}

@misc{nsw2024,
  author       = {{New South Wales Government}},
  title        = {COVID-19 Cases by Location},
  howpublished = {\url{https://data.nsw.gov.au/data/dataset/covid-19-cases-by-location}}
}

@misc{sciensanoCovid,
  author       = {{Sciensano}},
  title        = {COVID-19 Epidemiological Dashboard for Belgium},
  howpublished = {\url{https://epistat.sciensano.be/covid/}}
}

@misc{coronavirusBrazil,
  title        = {Corona Virus Brazil Dataset},
  howpublished = {\url{https://www.kaggle.com/datasets/unanimad/corona-virus-brazil}}
}

@misc{canadaCovid19,
  author       = {{Government of Canada}},
  title        = {COVID-19 Epidemiological Updates in Canada},
  howpublished = {\url{https://health-infobase.canada.ca/covid-19/}}
}

@misc{silvaAllendeCovid19,
  author       = {Alonso Silva Allende},
  title        = {COVID-19 Data Repository},
  howpublished = {\url{https://github.com/alonsosilvaallende/COVID-19/tree/master/data}}
}

@misc{colombiaCovid19,
  author       = {CovidDataProject},
  title        = {COVID-19 Data for Colombia},
  howpublished = {\url{https://github.com/CovidDataProject/DataCovid19Colombia}}
}

@misc{czechCovid19,
  author       = {Czech Ministry of Health},
  title        = {COVID-19 API for the Czech Republic},
  howpublished = {\url{https://onemocneni-aktualne.mzcr.cz/api/v2/covid-19}}
}

@misc{ecdcCovid19,
  author       = {European Centre for Disease Prevention and Control (ECDC)},
  title        = {Data on daily new cases of COVID-19 in EU/EEA by country},
  howpublished = {\url{https://www.ecdc.europa.eu/en/publications-data/data-daily-new-cases-covid-19-eueea-country}}
}

@misc{germanyCovid19,
  author       = {Jannis Gehrcke},
  title        = {COVID-19 Data for Germany},
  howpublished = {\url{https://github.com/jgehrcke/covid-19-germany-gae}}
}

@misc{indiaCovid19,
  author       = {India COVID-19 Data Project},
  title        = {COVID-19 Data for India},
  howpublished = {\url{https://data.incovid19.org}}
}

@misc{italyCovid19,
  author       = {Dipartimento della Protezione Civile},
  title        = {COVID-19 Data for Italian Provinces},
  howpublished = {\url{https://github.com/pcm-dpc/COVID-19/tree/master/dati-province}}
}

@misc{japanCovid19,
  author       = {Simon Fraser University DB Lab},
  title        = {COVID-19 Data for Japan},
  howpublished = {\url{https://github.com/sfu-db/covid19-datasets/blob/master/datasets-details/Japan.md}}
}

@misc{mexicoCovid19,
  author       = {CONACYT},
  title        = {COVID-19 Data for Mexico},
  howpublished = {\url{https://datos.covid-19.conacyt.mx}}
}

@misc{netherlandsCovid19,
  author       = {RIVM (Netherlands)},
  title        = {COVID-19 Data for the Netherlands},
  howpublished = {\url{https://data.rivm.nl/covid-19}}
}

@misc{spainCovid19,
  author       = {Victor Vicente Palacios},
  title        = {COVID-19 Data for Spain},
  howpublished = {\url{https://github.com/victorvicpal}}
}

@misc{switzerlandCovid19,
  author       = {Swiss Federal Office of Public Health (FOPH)},
  title        = {COVID-19 Data for Switzerland},
  howpublished = {\url{https://idd.bag.admin.ch/diseases/covid/data}}
}

@misc{ukCovid19,
  author       = {Tom White},
  title        = {COVID-19 Data for the United Kingdom},
  howpublished = {\url{https://github.com/tomwhite/covid-19-uk-data}}
}

@misc{irelandCovid19,
  author       = {Ireland Data Repository},
  title        = {COVID-19 Data for Ireland},
  howpublished = {\url{https://respiratoryvirus.hpsc.ie/pages/covid-19}}
}

@misc{usCovid19,
  author       = {The New York Times},
  title        = {COVID-19 Data for the United States},
  howpublished = {\url{https://github.com/nytimes/covid-19-data}}
}

@article{clarke2024,
  author       = {Clarke, J. and Lim, A. and Gupte, P. and Pigott, D.M. and van Panhuis, W.G. and Brady, O.J.},
  title        = {A global dataset of publicly available dengue case count data},
  journal      = {Scientific Data},
  volume       = {11},
  number       = {1},
  pages        = {296},
  year         = {2024},
  month        = {Mar 14},
  doi          = {10.1038/s41597-024-02703-8}
}

@article{lowe2021,
  author       = {Lowe, R. and Lee, S.A. and O'Reilly, K.M. and Brady, O.J. and Bastos, L. and Carrasco-Escobar, G. and de Castro Catão, R. and Colón-González, F.J. and Barcellos, C. and Carvalho, M.S. and Blangiardo, M. and Rue, H. and Gasparrini, A.},
  title        = {Combined effects of hydrometeorological hazards and urbanisation on dengue risk in Brazil: a spatiotemporal modelling study},
  journal      = {The Lancet Planetary Health},
  volume       = {5},
  number       = {4},
  pages        = {e209--e219},
  year         = {2021},
  month        = {Apr},
  doi          = {10.1016/S2542-5196(21)00027-6}
}

@misc{rDengueAnalysis,
  author       = {Hassan, Shak},
  title        = {R Dengue Analysis},
  howpublished = {\url{https://github.com/shakhassan/r-dengue-analysis}}
}

@misc{pahangWDF,
  author       = {Ping, Foo},
  title        = {Pahang WDF: Weather-Dengue Forecasting},
  howpublished = {\url{https://github.com/ping543f/pahang-wdf/tree/main}}
}

@article{morris2023,
  author       = {Morris, M. and Hayes, P. and Cox, I.J. and Lampos, V.},
  title        = {Neural network models for influenza forecasting with associated uncertainty using Web search activity trends},
  journal      = {PLoS Computational Biology},
  volume       = {19},
  number       = {8},
  pages        = {e1011392},
  year         = {2023},
  doi          = {10.1371/journal.pcbi.1011392}
}

@misc{projectTycho,
  author       = {Project Tycho, University of Pittsburgh},
  title        = {Project Tycho Data Repository},
  howpublished = {\url{https://www.tycho.pitt.edu/}}
}

@misc{cdcepiZika,
  author       = {Centers for Disease Control and Prevention (CDC)},
  title        = {Zika Virus Data Repository},
  howpublished = {\url{https://github.com/cdcepi/zika}}
}

@article{colon2023projecting,
  title={Projecting the future incidence and burden of dengue in Southeast Asia},
  author={Col{\'o}n-Gonz{\'a}lez, Felipe J and Gibb, Rory and Khan, Kamran and Watts, Alexander and Lowe, Rachel and Brady, Oliver J},
  journal={nature communications},
  volume={14},
  number={1},
  pages={5439},
  year={2023},
  publisher={Nature Publishing Group UK London}
}

@article{barman2025epidemic,
  title={Epidemic-guided deep learning for spatiotemporal forecasting of tuberculosis outbreak},
  author={Barman, Madhab and Panja, Madhurima and Mishra, Nachiketa and Chakraborty, Tanujit},
  journal={Machine Learning},
  volume={114},
  number={10},
  pages={213},
  year={2025},
  publisher={Springer}
}

@article{breiman2001random,
  title={Random forests},
  author={Breiman, Leo},
  journal={Machine learning},
  volume={45},
  number={1},
  pages={5--32},
  year={2001},
  publisher={Springer}
}

@inproceedings{chen2016xgboost,
  title={Xgboost: A scalable tree boosting system},
  author={Chen, Tianqi and Guestrin, Carlos},
  booktitle={Proceedings of the 22nd ACM SIGKDD International Conference on Knowledge Discovery and Data Mining},
  pages={785--794},
  year={2016}
}

@inproceedings{challu2023nhits,
  title={Nhits: Neural hierarchical interpolation for time series forecasting},
  author={Challu, Cristian and Olivares, Kin G and Oreshkin, Boris N and Ramirez, Federico Garza and Canseco, Max Mergenthaler and Dubrawski, Artur},
  booktitle={Proceedings of the AAAI conference on artificial intelligence},
  volume={37},
  pages={6989--6997},
  year={2023}
}

@article{das2023longterm,
title={Long-term Forecasting with Ti{DE}: Time-series Dense Encoder},
author={Abhimanyu Das and Weihao Kong and Andrew Leach and Shaan K Mathur and Rajat Sen and Rose Yu},
journal={Transactions on Machine Learning Research},
issn={2835-8856},
year={2023},
}

@inproceedings{oreshkin2020nbeats,
title={N-BEATS: Neural basis expansion analysis for interpretable time series forecasting},
author={Boris N. Oreshkin and Dmitri Carpov and Nicolas Chapados and Yoshua Bengio},
booktitle={International Conference on Learning Representations},
year={2020}
}

@article{salinas2020deepar,
  title={DeepAR: Probabilistic forecasting with autoregressive recurrent networks},
  author={Salinas, David and Flunkert, Valentin and Gasthaus, Jan and Januschowski, Tim},
  journal={International Journal of Forecasting},
  volume={36},
  number={3},
  pages={1181--1191},
  year={2020},
  publisher={Elsevier}
}

@inproceedings{zeng2023transformers,
  title={Are transformers effective for time series forecasting?},
  author={Zeng, Ailing and Chen, Muxi and Zhang, Lei and Xu, Qiang},
  booktitle={Proceedings of the AAAI conference on artificial intelligence},
  volume={37},
  pages={11121--11128},
  year={2023}
}

@article{chen2023tsmixer,
title={{TSM}ixer: An All-{MLP} Architecture for Time Series Forecast-ing},
author={Si-An Chen and Chun-Liang Li and Sercan O Arik and Nathanael Christian Yoder and Tomas Pfister},
journal={Transactions on Machine Learning Research},
issn={2835-8856},
year={2023}
}

@inproceedings{liu2025kan,
title={{KAN}: Kolmogorov{\textendash}Arnold Networks},
author={Ziming Liu and Yixuan Wang and Sachin Vaidya and Fabian Ruehle and James Halverson and Marin Soljacic and Thomas Y. Hou and Max Tegmark},
booktitle={The Thirteenth International Conference on Learning Representations},
year={2025}
}

@article{hochreiter1997long,
  title={Long short-term memory},
  author={Hochreiter, Sepp and Schmidhuber, J{\"u}rgen},
  journal={Neural computation},
  volume={9},
  number={8},
  pages={1735--1780},
  year={1997},
  publisher={MIT press}
}

@article{herzen2022darts,
  title={Darts: User-friendly modern machine learning for time series},
  author={Herzen, Julien and L{\"a}ssig, Francesco and Piazzetta, Samuele Giuliano and Neuer, Thomas and Tafti, L{\'e}o and Raille, Guillaume and Van Pottelbergh, Tomas and Pasieka, Marek and Skrodzki, Andrzej and Huguenin, Nicolas and others},
  journal={Journal of Machine Learning Research},
  volume={23},
  number={124},
  pages={1--6},
  year={2022}
}

@article{chen2020probabilistic,
  title={Probabilistic forecasting with temporal convolutional neural network},
  author={Chen, Yitian and Kang, Yanfei and Chen, Yixiong and Wang, Zizhuo},
  journal={Neurocomputing},
  volume={399},
  pages={491--501},
  year={2020},
  publisher={Elsevier}
}

@article{vaswani2017attention,
  title={Attention is all you need},
  author={Vaswani, Ashish and Shazeer, Noam and Parmar, Niki and Uszkoreit, Jakob and Jones, Llion and Gomez, Aidan N and Kaiser, {\L}ukasz and Polosukhin, Illia},
  journal={Advances in neural information processing systems},
  volume={30},
  year={2017}
}

@article{ansari2025chronos,
  title={Chronos-2: From univariate to universal forecasting},
  author={Ansari, Abdul Fatir and Shchur, Oleksandr and K{\"u}ken, Jaris and Auer, Andreas and Han, Boran and Mercado, Pedro and Rangapuram, Syama Sundar and Shen, Huibin and Stella, Lorenzo and Zhang, Xiyuan and others},
  journal={arXiv preprint arXiv:2510.15821},
  year={2025}
}

@inproceedings{das2024decoder,
  title={A decoder-only foundation model for time-series forecasting},
  author={Das, Abhimanyu and Kong, Weihao and Sen, Rajat and Zhou, Yichen},
  booktitle={Forty-first International Conference on Machine Learning},
  year={2024}
}

@inproceedings{godahewa2021monash,
  title={Monash time series forecasting archive},
  author={Godahewa, Rakshitha and Bergmeir, Christoph and Webb, Geoffrey I and Hyndman, Rob J and Montero-Manso, Pablo},
  booktitle={35th Conference on Neural Information Processing Systems (NeurIPS 2021) Track on Datasets and Benchmarks},
  year={2021}
}

@article{hyndman2006another,
  title={Another look at measures of forecast accuracy},
  author={Hyndman, Rob J and Koehler, Anne B},
  journal={International Journal of Forecasting},
  volume={22},
  number={4},
  pages={679--688},
  year={2006},
  publisher={Elsevier}
}

@article{panja2023epicasting,
  title={Epicasting: an ensemble wavelet neural network for forecasting epidemics},
  author={Panja, Madhurima and Chakraborty, Tanujit and Kumar, Uttam and Liu, Nan},
  journal={Neural Networks},
  volume={165},
  pages={185--212},
  year={2023},
  publisher={Elsevier}
}

@article{koning2005m3,
  title={The M3 competition: Statistical tests of the results},
  author={Koning, Alex J and Franses, Philip Hans and Hibon, Mich{\`e}le and Stekler, Herman O},
  journal={International Journal of Forecasting},
  volume={21},
  number={3},
  pages={397--409},
  year={2005},
  publisher={Elsevier}
}

@article{friedman1937use,
  title={The use of ranks to avoid the assumption of normality implicit in the analysis of variance},
  author={Friedman, Milton},
  journal={Journal of the American Statistical Association},
  volume={32},
  number={200},
  pages={675--701},
  year={1937},
  publisher={Taylor \& Francis}
}

@article{friedman1940comparison,
  title={A comparison of alternative tests of significance for the problem of m rankings},
  author={Friedman, Milton},
  journal={The Annals of Mathematical Statistics},
  volume={11},
  number={1},
  pages={86--92},
  year={1940},
  publisher={JSTOR}
}

@article{lv2021time,
  title={Time series analysis of hemorrhagic fever with renal syndrome in mainland China by using an XGBoost forecasting model},
  author={Lv, Cai-Xia and An, Shu-Yi and Qiao, Bao-Jun and Wu, Wei},
  journal={BMC infectious diseases},
  volume={21},
  number={1},
  pages={839},
  year={2021},
  publisher={Springer}
}

@article{chimmula2020time,
  title={Time series forecasting of COVID-19 transmission in Canada using LSTM networks},
  author={Chimmula, Vinay Kumar Reddy and Zhang, Lei},
  journal={Chaos, solitons \& fractals},
  volume={135},
  pages={109864},
  year={2020},
  publisher={Elsevier}
}

@article{rodriguez2024machine,
  title={Machine learning for data-centric epidemic forecasting},
  author={Rodriguez, Alexander and Kamarthi, Harshavardhan and Agarwal, Pulak and Ho, Javen and Patel, Mira and Sapre, Suchet and Prakash, B Aditya},
  journal={Nature Machine Intelligence},
  volume={6},
  number={10},
  pages={1122--1131},
  year={2024},
  publisher={Nature Publishing Group UK London}
}

@article{wu2020deep,
  title={Deep transformer models for time series forecasting: The influenza prevalence case},
  author={Wu, Neo and Green, Bradley and Ben, Xue and O'Banion, Shawn},
  journal={arXiv preprint arXiv:2001.08317},
  year={2020}
}

@article{gong2025epillm,
  title={EpiLLM: unlocking the potential of large language models in epidemic forecasting},
  author={Gong, Chenghua and Sun, Rui and Zheng, Yuhao and Zhang, Juyuan and Gu, Tianjun and Pan, Liming and Lv, Linyuan},
  journal={arXiv preprint arXiv:2505.12738},
  year={2025}
}

@article{adiga2026idobe,
  title={IDOBE: Infectious Disease Outbreak forecasting Benchmark Ecosystem},
  author={Adiga, Aniruddha and Chou, Jingyuan and Chiranth, Anshul and Lewis, Bryan and Bento, Ana I and Truelove, Shaun and Fox, Geoffrey and Marathe, Madhav and Hochheiser, Harry and Venkatramanan, Srini},
  journal={arXiv preprint arXiv:2604.18521},
  year={2026}
}

@article{biggerstaff2016results,
  title={Results from the centers for disease control and prevention’s predict the 2013--2014 Influenza Season Challenge},
  author={Biggerstaff, Matthew and Alper, David and Dredze, Mark and Fox, Spencer and Fung, Isaac Chun-Hai and Hickmann, Kyle S and Lewis, Bryan and Rosenfeld, Roni and Shaman, Jeffrey and Tsou, Ming-Hsiang and others},
  journal={BMC infectious diseases},
  volume={16},
  number={1},
  pages={357},
  year={2016},
  publisher={Springer}
}

@article{johansson2019open,
  title={An open challenge to advance probabilistic forecasting for dengue epidemics},
  author={Johansson, Michael A and Apfeldorf, Karyn M and Dobson, Scott and Devita, Jason and Buczak, Anna L and Baugher, Benjamin and Moniz, Linda J and Bagley, Thomas and Babin, Steven M and Guven, Erhan and others},
  journal={Proceedings of the National Academy of Sciences},
  volume={116},
  number={48},
  pages={24268--24274},
  year={2019},
  publisher={National Academy of Sciences}
}

@article{cramer2022evaluation,
  title={Evaluation of individual and ensemble probabilistic forecasts of COVID-19 mortality in the United States},
  author={Cramer, Estee Y and Ray, Evan L and Lopez, Velma K and Bracher, Johannes and Brennen, Andrea and Castro Rivadeneira, Alvaro J and Gerding, Aaron and Gneiting, Tilmann and House, Katie H and Huang, Yuxin and others},
  journal={Proceedings of the National Academy of Sciences},
  volume={119},
  number={15},
  pages={e2113561119},
  year={2022},
  publisher={National Academy of Sciences}
}

@article{goehry2023random,
  title={Random forests for time series},
  author={Goehry, Benjamin and Yan, Hui and Goude, Yannig and Massart, Pascal and Poggi, Jean-Michel},
  journal={REVSTAT-Statistical Journal},
  volume={21},
  number={2},
  pages={283--302},
  year={2023}
}

@article{brooks2018nonmechanistic,
  title={Nonmechanistic forecasts of seasonal influenza with iterative one-week-ahead distributions},
  author={Brooks, Logan C and Farrow, David C and Hyun, Sangwon and Tibshirani, Ryan J and Rosenfeld, Roni},
  journal={PLoS computational biology},
  volume={14},
  number={6},
  pages={e1006134},
  year={2018},
  publisher={Public Library of Science San Francisco, CA USA}
}

@article{rosenfeld2021epidemic,
  title={Epidemic tracking and forecasting: Lessons learned from a tumultuous year},
  author={Rosenfeld, Roni and Tibshirani, Ryan J},
  journal={Proceedings of the National Academy of Sciences},
  volume={118},
  number={51},
  pages={e2111456118},
  year={2021},
  publisher={National Academy of Sciences}
}

@misc{hyndman2023tsfeatures,
  title={tsfeatures: Time Series Feature Extraction},
  author={Hyndman, Rob and Kang, Yanfei and Montero-Manso, Pablo and O'Hara-Wild, Mitchell and Talagala, Thiyanga and Wang, Earo and Yang, Yangzhuoran and Taieb, Souhaib Ben and Hanqing, Cao and Lake, D. K. and Laptev, Nikolay and Moorman, J. R. and Zhang, Bohan},
  year={2023}
}



\clearpage

\appendix




\appendix
\section{Appendix}

\subsection{Baseline models}
In EpiCastBench, we benchmark the performance of fifteen forecasting models of different paradigms. A brief description of these models is outlined as follows:
\begin{itemize}
    \item \emph{Naive} (or Random Walk) model is among the most fundamental statistical baseline architectures, in which the value of a time series at each step evolves as a random deviation from its previous value.
    \item \emph{DLinear} model is a variant of linear forecasting approaches that initially decomposes the input time series into distinct components and then models them independently \cite{zeng2023transformers}. Specifically, a moving average operation is used to capture the trend component, while the residual part is treated as the seasonal component. Each component is passed through its own linear layer, and their outputs are subsequently combined to produce the final forecast. By explicitly disentangling long-term trends from seasonal fluctuations, this framework maintains simplicity in its architecture while achieving competitive forecasting performance. 
    \item \emph{Random Forest} is a supervised learning technique that builds a bagging-based ensemble of decision trees, with each tree being trained on a different resampled version of the data \cite{breiman2001random}. For time series forecasting, a blocked bootstrap strategy is often adopted, in which consecutive segments of the series are sampled to preserve temporal dependencies \cite{goehry2023random}. The final prediction is obtained by aggregating the outputs of all trees, typically through averaging, which helps reduce variance and improve stability. Due to its ability to model nonlinear relationships and interactions within sequential data, Random Forest serves as a critical baseline in forecasting applications.
    \item \emph{XGBoost} is an advanced ensemble learning method that enhances gradient-boosted decision trees through optimization and parallel computation \cite{chen2016xgboost}. Given an input series, the model builds a sequence of decision trees, with each new tree focusing on correcting the errors made by the previous ones. To prevent overfitting, XGBoost incorporates regularization that controls model complexity. Through this iterative refinement process, it can capture complex nonlinear patterns and interactions in the data, making it a powerful approach for modeling time series with intricate temporal dependencies.
    \item \emph{TSMixer} is a multi-layer perceptron (MLP)-based time series forecasting approach that leverages a sequence-mixing mechanism to capture relationships across time steps \cite{chen2023tsmixer}. The initial layers focus on aggregating sequential patterns, while deeper layers iteratively refine these representations to improve predictive performance. This approach allows the model to effectively learn complex time-based interactions while maintaining a simple and computationally efficient structure.    
    \item \emph{KAN}, recently developed based on the Kolmogorov–Arnold representation theorem, offers an alternative to conventional MLP architectures \cite{liu2025kan}. It replaces fixed neuron activation functions with learnable nonlinear transformations on connections, where standard weights are represented using parameterized univariate spline functions. This flexible formulation enables the model to better capture complex nonlinear relationships in the data. For time series forecasting, such a design makes KANs well-suited for modeling intricate temporal dynamics. 
    \item \emph{LSTM} networks are a specialized form of recurrent neural networks (RNNs) designed to effectively learn long-term dependencies in sequential data \cite{hochreiter1997long}. This framework incorporates a structured gating mechanism to mitigate the vanishing and exploding gradients problems of conventional RNNs. The training mechanism of LSTM is regulated by forget, input, and output gates, which control the amount of information to be retained, updated, or removed over time. This design allows the model to selectively preserve critical historical information while filtering out irrelevant details. As a result, LSTMs are particularly well-suited for time series forecasting tasks where patterns from the distant past can significantly influence future values.
    \item \emph{DeepAR} framework incorporates modified LSTM memory cells to learn inherent patterns in time series data \citep{salinas2020deepar}. It adopts an encoder-decoder framework that enables the model to better capture complex nonlinear structures and long-range relationships compared to standard LSTM architectures. This design makes DeepAR particularly effective for forecasting tasks characterized by limited data availability and high variability.
    \item \emph{TCN} is a convolutional neural network-based approach designed for time series forecasting \cite{chen2020probabilistic}. It employs causal convolutions to ensure that predictions at any time step depend only on historical inputs, while dilated convolutions enable efficient learning of long-range dependencies without requiring deep architectures. In addition, residual connections are used to stabilize training and improve gradient propagation, making the model easier to optimize for sequential learning tasks.
    \item \emph{NBeats} is a deep learning-based time series forecasting technique \cite{oreshkin2020nbeats}. It consists of a fully connected feedforward network arranged as a stack of sequential blocks, where each block progressively refines the prediction. Within each block, the input series is decomposed into two components: one reconstructs the observed signal, while the other contributes to the forecast. The model is trained in a recursive manner, where successive blocks focus on correcting the remaining errors from previous blocks. This structural refinement enables NBeats to learn both long-term patterns and short-term variations in time series data, making it an effective benchmark across a wide range of forecasting tasks.
    \item \emph{NHiTS} extends the NBeats framework by introducing a hierarchical, multi-scale architecture for time series forecasting \cite{challu2023nhits}. The model is structured into a sequence of blocks that first compress the input series to capture coarse, low-frequency patterns. These representations are then progressively reconstructed to the original resolution using learned upsampling techniques. This multi-resolution design enables the model to effectively represent both long-term trends and short-term fluctuations across different temporal scales. Furthermore, residual connections between blocks allow iterative correction of forecasting errors, leading to improved performance, particularly for long-horizon predictions.
    \item \emph{Transformers} represent the state-of-the-art self-attention-based technique that model complex dependencies in sequential data \cite{vaswani2017attention}. The training mechanism of Transformers allows the model to process the entire input sequence simultaneously, which improves computational efficiency and scalability. Through multi-head attention, the model can focus on different parts of the input at the same time, enabling it to learn complex relationships and interactions across different time steps. This makes Transformer-based architectures well-suited for capturing complex, non-linear, and multi-scale temporal patterns in forecasting problems.
    \item \emph{TiDE} is an encoder–decoder-based architecture that models time series data using MLPs \cite{das2023longterm}. The encoder converts historical observations into a compact latent representation using fully connected projection layers, while the decoder translates this representation into future predictions. Both components are built using residual blocks with multiple hidden layers and skip connections, which help in capturing complex and non-linear temporal patterns more effectively. 
    \item \emph{Chronos-2} is a pretrained multivariate time series forecasting technique designed to perform inference without task-specific training \cite{ansari2025chronos}. This zero-shot prediction framework employs attention-based in-context learning for efficient information retrieval across temporal sequences. The model has been pretrained on large-scale synthetic datasets, resulting in better generalization for real-world forecasting tasks. 
    \item \emph{TimesFM} is a decoder-based foundation model designed for time series forecasting tasks \cite{das2024decoder}. This pretrained framework employs a decoder-style attention mechanism with input patching on a large and diverse corpus of both real-world and synthetic time series data. This pretraining strategy enables the model to learn generalizable temporal patterns across domains, forecasting horizons, and sampling frequencies. 
    
\end{itemize}
 
\subsection{Performance evaluation measures}
The selection of appropriate error metrics for forecasting remains challenging, particularly for epidemic time series characterized by varying trends, fluctuations, and zero-inflated counts. Many conventional performance measures can yield misleading evaluations depending on the underlying data properties. To ensure a robust assessment in EpiCastBench, we evaluate model performance using multiple widely adopted metrics, providing complementary perspectives across diverse epidemic settings \cite{hyndman2006another, panja2023epicasting}. Specifically, we consider scale-independent MASE and SMAPE measures, along with scale-dependent metrics such as MAE and RMSE. The mathematical definitions of these metrics are given by
\begin{align*}
    \text{ MASE} & = \frac{\sum_{t = N + 1}^{N+h} |Y_t - \widehat{Y}_t|}{\frac{h}{N-1} \sum_{t = 2}^N |Y_t - Y_{t-1}|}, \quad \text{ SMAPE} = \frac{1}{h} \sum_{t=1}^h \frac{2|Y_t - \widehat{Y}_t|}{|Y_t| + |\widehat{Y}_t|} \times 100 \%, \\
    \text{ MAE} & = \frac{1}{h}\sum_{t =1}^{h} |Y_t - \widehat{Y}_t|, \quad \text{and} \quad  \text{ RMSE} = \sqrt{\frac{1}{h}\sum_{t =1}^{h} (Y_t - \widehat{Y}_t)^2},
\end{align*}
where $h$ denotes the forecast horizon, $\widehat{Y}_t$ is the forecast of the actual value $Y_t$ at time $t$, and $N$ is the size of the training sample. For each multivariate dataset, metrics are computed per time series and then averaged across all series to obtain the final reported performance.

\begin{figure}
    \centering
    \includegraphics[width=1\linewidth]{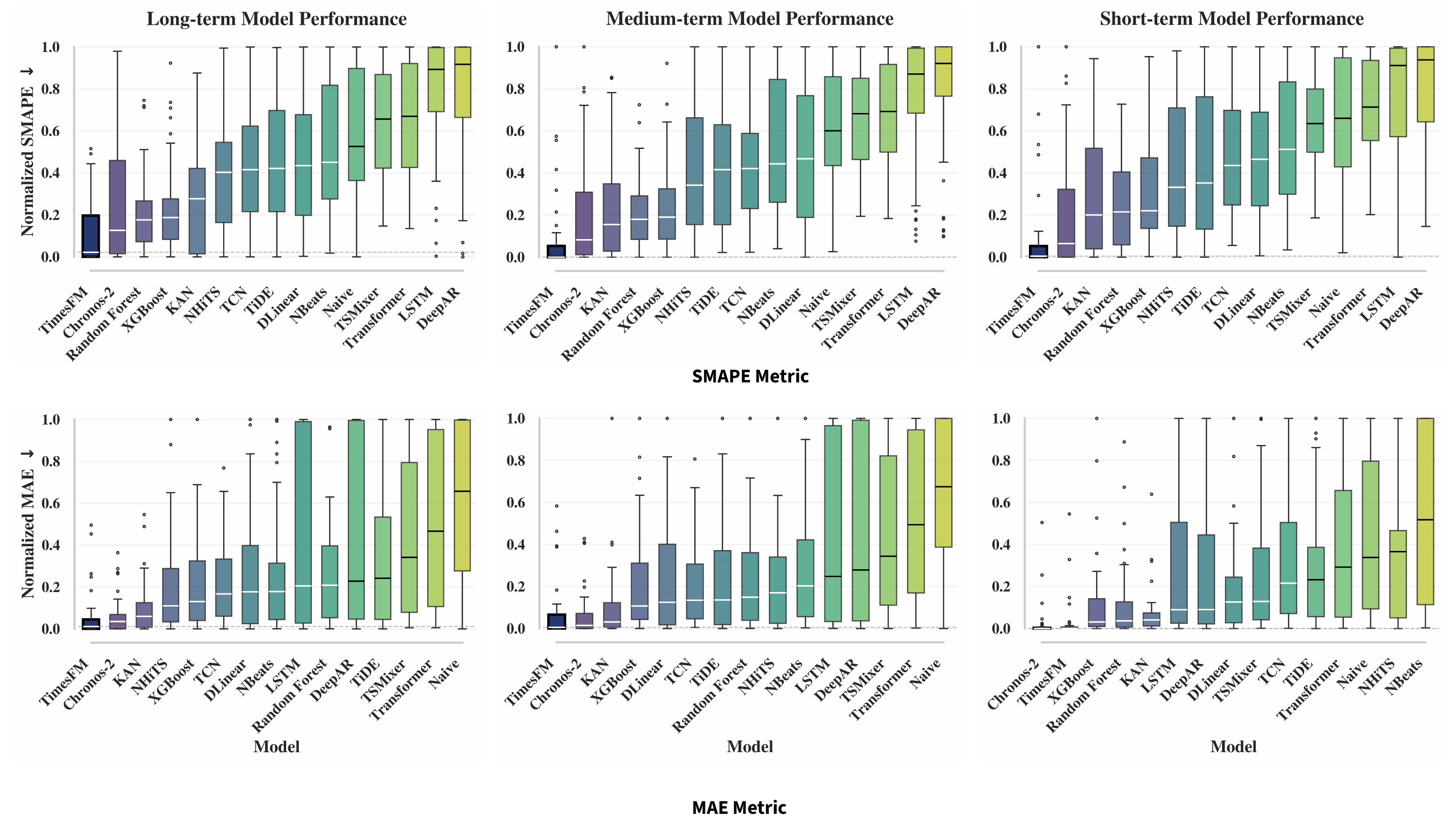}
    \caption{Boxplots comparing the performance of different models based on SMAPE (upper panel) and MAE (lower panel) across long (left), medium (middle), and short-term (right) forecasting horizons.}
    \label{Fig_Boxplot_SMAPE_MAE}
\end{figure}

\begin{table}[]
    \centering
    \caption{Short-term forecasting performance of baseline and state-of-the-art models evaluated using the MASE metric. The best results are indicated in \underline{\textbf{bold}}, while the second-best results are \textit{highlighted}. The average rank of each model for the specific forecasting task is reported at the end of the table.}
    \fontsize{5.9}{7}\selectfont
    \setlength{\tabcolsep}{0.7pt}
\begin{tabular}{cccccccccccccccccccccc}
    \toprule
    \multirow{2}{*}{Disease}	&	\multirow{2}{*}{Region}	&	\multirow{2}{*}{Naive}	&	\multirow{2}{*}{DLinear}	&	Random	&	\multirow{2}{*}{XGBoost}	&	\multirow{2}{*}{TSMixer}	&	\multirow{2}{*}{KAN}	&	\multirow{2}{*}{LSTM}	&	\multirow{2}{*}{DeepAR}	&	\multirow{2}{*}{TCN}	&	\multirow{2}{*}{NBeats}	&	\multirow{2}{*}{NHiTS}	&	Trans-	&	\multirow{2}{*}{TiDE}	&	\multirow{2}{*}{Chronos-2}	&	\multirow{2}{*}{TimesFM}	\\
    &	&	&	& Forest & & & & &	&	& 	&	&	former	&	&	&	\\  \midrule

    \multirow{21}{*}{Covid}	&	Australia	&	3.886	&	0.831	&	0.179	&	0.162	&	0.341	&	0.310	&	0.621	&	0.728	&	0.292	&	1.960	&	0.584	&	6.46	&	0.595	&	\underline{\textbf{0.101}}	&	\textit{0.104}	\\
	&	Belgium	&	2.375	&	0.038	&	0.029	&	0.038	&	0.156	&	0.057	&	0.097	&	0.095	&	0.095	&	0.049	&	0.043	&	3.071	&	0.033	&	\underline{\textbf{0.025}}	&	\textit{0.026}	\\
	&	Brazil	&	1.880	&	1.525	&	1.219	&	1.062	&	3.378	&	0.999	&	3.752	&	3.737	&	1.454	&	1.662	&	1.439	&	3.369	&	1.223	&	\textit{0.858}	&	\underline{\textbf{0.856}}	\\
	&	Canada	&	6.523	&	72.463	&	3.844	&	5.571	&	7.180	&	7.502	&	7.317	&	7.345	&	616.196	&	1111.698	&	512.292	&	7.611	&	371.393	&	\textit{2.425}	&	\underline{\textbf{2.401}}	\\
	&	Canada 2	&	3.497	&	0.658	&	0.198	&	0.357	&	0.673	&	0.241	&	0.381	&	0.391	&	0.750	&	8.637	&	3.233	&	0.893	&	3.260	&	\underline{\textbf{0.134}}	&	\textit{0.151}	\\
	&	Chile	&	\textit{1.165}	&	12.720	&	2.942	&	3.533	&	1.244	&	3.022	&	2.030	&	\underline{\textbf{1.131}}	&	12.030	&	24.242	&	20.487	&	1.507	&	18.378	&	1.305	&	1.834	\\
\bottomrule
        \end{tabular}
    \label{table_mase_short}
\end{table}

\begin{table}[]
    \centering
    \caption{Short-term forecasting performance of baseline and state-of-the-art models evaluated using the RMSE metric. The best results are indicated in \underline{\textbf{bold}}, while the second-best results are \textit{highlighted}. The average rank of each model for the specific forecasting task is reported at the end of the table.}
    \fontsize{5.5}{7}\selectfont
    \setlength{\tabcolsep}{0.7pt}
    \begin{tabular}{cccccccccccccccccccccc}
    \toprule
    \multirow{2}{*}{Disease}	&	\multirow{2}{*}{Region}	&	\multirow{2}{*}{Naive}	&	\multirow{2}{*}{DLinear}	&	Random	&	\multirow{2}{*}{XGBoost}	&	\multirow{2}{*}{TSMixer}	&	\multirow{2}{*}{KAN}	&	\multirow{2}{*}{LSTM}	&	\multirow{2}{*}{DeepAR}	&	\multirow{2}{*}{TCN}	&	\multirow{2}{*}{NBeats}	&	\multirow{2}{*}{NHiTS}	&	Trans-	&	\multirow{2}{*}{TiDE}	&	\multirow{2}{*}{Chronos-2}	&	\multirow{2}{*}{TimesFM}	\\
    &	&	&	& Forest & & & & &	&	& 	&	&	former	&	&	&	\\  \midrule
\multirow{21}{*}{Covid}	&	Australia	&	160.686	&	10.776	&	7.211	&	6.819	&	11.550	&	5.706	&	13.862	&	14.850	&	7.759	&	6.717	&	6.071	&	175.617	&	5.398	&	\textit{4.645}	&	\underline{\textbf{4.644}}	\\
	&	Belgium	&	291.39	&	5.488	&	\textit{4.656}	&	6.509	&	11.410	&	8.804	&	9.412	&	8.970	&	14.213	&	6.219	&	5.520	&	292.901	&	5.099	&	4.839	&	\underline{\textbf{4.405}}	\\
	&	Brazil	&	1435.83	&	979.206	&	769.714	&	694.601	&	2466.522	&	626.426	&	2578.459	&	2574.306	&	949.77	&	890.851	&	912.744	&	2468.963	&	834.473	&	\underline{\textbf{557.223}}	&	\textit{571.514}	\\
	&	Canada	&	346.411	&	113.433	&	151.423	&	147.859	&	397.651	&	98.528	&	414.725	&	414.757	&	174.513	&	173.518	&	124.173	&	397.670	&	134.161	&	\underline{\textbf{79.417}}	&	\textit{81.147}	\\
	&	Canada 2	&	1561.296	&	116.613	&	102.496	&	169.730	&	161.130	&	82.056	&	166.823	&	166.501	&	92.643	&	622.368	&	363.696	&	161.158	&	348.942	&	\underline{\textbf{56.308}}	&	\textit{63.838}	\\
	&	Chile	&	2.370	&	13.640	&	8.394	&	8.965	&	\underline{\textbf{1.908}}	&	6.358	&	2.094	&	\textit{1.959}	&	7.085	&	11.925	&	12.868	&	1.964	&	11.558	&	3.466	&	6.315	\\
\bottomrule       
    \end{tabular}
    \label{table_rmse_short}
\end{table}

\begin{table}[]
    \centering
    \caption{Short-term forecasting performance of baseline and state-of-the-art models evaluated using the MAE metric. The best results are indicated in \underline{\textbf{bold}}, while the second-best results are \textit{highlighted}. The average rank of each model for the specific forecasting task is reported at the end of the table.}
    \fontsize{5.5}{7}\selectfont
    \setlength{\tabcolsep}{0.5pt}
    \begin{tabular}{cccccccccccccccccccccc}
    \toprule
    \multirow{2}{*}{Disease}	&	\multirow{2}{*}{Region}	&	\multirow{2}{*}{Naive}	&	\multirow{2}{*}{DLinear}	&	Random	&	\multirow{2}{*}{XGBoost}	&	\multirow{2}{*}{TSMixer}	&	\multirow{2}{*}{KAN}	&	\multirow{2}{*}{LSTM}	&	\multirow{2}{*}{DeepAR}	&	\multirow{2}{*}{TCN}	&	\multirow{2}{*}{NBeats}	&	\multirow{2}{*}{NHiTS}	&	Trans-	&	\multirow{2}{*}{TiDE}	&	\multirow{2}{*}{Chronos-2}	&	\multirow{2}{*}{TimesFM}	\\
    &	&	&	& Forest & & & & &	&	& 	&	&	former	&	&	&	\\  \midrule

    \multirow{21}{*}{Covid}	&	Australia	&	160.590	&	9.755	&	6.418	&	5.673	&	9.735	&	4.674	&	12.506	&	13.844	&	6.492	&	5.824	&	4.787	&	257.555	&	4.307	&	\underline{\textbf{3.779}}	&	\textit{3.840}	\\
	&	Belgium	&	291.356	&	4.594	&	3.781	&	5.209	&	10.289	&	6.831	&	8.475	&	8.308	&	11.503	&	5.020	&	4.684	&	362.871	&	3.985	&	\textit{3.678}	&	\underline{\textbf{3.524}}	\\
	&	Brazil	&	1266.634	&	824.633	&	649.717	&	570.688	&	2282.598	&	538.445	&	2397.900	&	2393.442	&	813.020	&	731.845	&	806.822	&	2280.244	&	676.395	&	\underline{\textbf{467.483}}	&	\textit{474.475}	\\
	&	Canada	&	329.606	&	97.202	&	134.374	&	132.627	&	380.829	&	82.633	&	396.934	&	396.984	&	150.083	&	151.431	&	100.506	&	380.743	&	113.715	&	\underline{\textbf{67.963}}	&	\textit{69.890}	\\
	&	Canada 2	&	1560.627	&	105.924	&	97.280	&	154.202	&	157.670	&	70.389	&	162.161	&	161.909	&	84.669	&	538.627	&	330.911	&	158.179	&	324.193	&	\underline{\textbf{48.866}}	&	\textit{58.219}	\\
	&	Chile	&	2.039	&	13.343	&	8.185	&	8.689	&	1.687	&	5.871	&	\textit{1.578}	&	\underline{\textbf{1.443}}	&	6.426	&	10.968	&	12.162	&	1.666	&	10.344	&	2.931	&	5.716	\\
\bottomrule
            \end{tabular}
    \label{table_mae_short}
\end{table}

\begin{table}[]
    \centering
    \caption{Short-term forecasting performance of baseline and state-of-the-art models evaluated using the SMAPE metric. The best results are indicated in \underline{\textbf{bold}}, while the second-best results are \textit{highlighted}. The average rank of each model for the specific forecasting task is reported at the end of the table.}
    \fontsize{6}{7}\selectfont
    \setlength{\tabcolsep}{0.7pt}
    \begin{tabular}{cccccccccccccccccccccc}
    \toprule
    \multirow{2}{*}{Disease}	&	\multirow{2}{*}{Region}	&	\multirow{2}{*}{Naive}	&	\multirow{2}{*}{DLinear}	&	Random	&	\multirow{2}{*}{XGBoost}	&	\multirow{2}{*}{TSMixer}	&	\multirow{2}{*}{KAN}	&	\multirow{2}{*}{LSTM}	&	\multirow{2}{*}{DeepAR}	&	\multirow{2}{*}{TCN}	&	\multirow{2}{*}{NBeats}	&	\multirow{2}{*}{NHiTS}	&	Trans-	&	\multirow{2}{*}{TiDE}	&	\multirow{2}{*}{Chronos-2}	&	\multirow{2}{*}{TimesFM}	\\
    &	&	&	& Forest & & & & &	&	& 	&	&	former	&	&	&	\\  \midrule
        \multirow{21}{*}{Covid}	&	Australia	&	177.713	&	129.886	&	76.130	&	77.822	&	137.552	&	75.293	&	102.651	&	106.507	&	81.319	&	77.645	&	72.451	&	185.694	&	72.213	&	\textit{69.591}	&	\underline{\textbf{44.576}}	\\
	&	Belgium	&	196.120	&	140.800	&	145.506	&	170.212	&	157.681	&	155.290	&	142.592	&	143.026	&	156.016	&	151.191	&	\textit{140.006}	&	196.857	&	143.127	&	151.018	&	\underline{\textbf{132.279}}	 \\
	&	Brazil	&	57.144	&	52.096	&	36.736	&	31.208	&	149.864	&	30.821	&	192.585	&	190.079	&	44.968	&	53.388	&	45.890	&	147.578	&	41.060	&	\textit{26.745}	&	\underline{\textbf{25.796}}	\\
	&	Canada	&	152.915	&	110.951	&	\underline{\textbf{71.018}}	&	96.259	&	162.125	&	99.055	&	166.960	&	171.986	&	123.135	&	133.300	&	124.084	&	163.226	&	123.512	&	99.887	&	\textit{73.349}	 \\
	&	Canada 2	&	175.899	&	161.819	&	\textit{76.841}	&	127.466	&	145.057	&	96.010	&	174.644	&	172.503	&	112.222	&	157.072	&	147.232	&	141.856	&	151.662	&	93.279	&	\underline{\textbf{47.861}}	\\
	&	Chile	&	173.364	&	180.228	&	\underline{\textbf{156.018}}	&	\textit{162.105}	&	176.269	&	175.381	&	174.216	&	176.527	&	185.977	&	190.519	&	183.239	&	175.212	&	187.506	&	180.694	&	179.488	 \\
\bottomrule

            \end{tabular}
    \label{table_smape_short}
\end{table}

\begin{table}[]
    \centering
    \caption{Medium-term forecasting performance of baseline and state-of-the-art models evaluated using the MASE metric. The best results are indicated in \underline{\textbf{bold}}, while the second-best results are \textit{highlighted}. The average rank of each model for the specific forecasting task is reported at the end of the table.}
    \fontsize{5.9}{7}\selectfont
    \setlength{\tabcolsep}{0.7pt}
    \begin{tabular}{cccccccccccccccccccccc}
    \toprule
    \multirow{2}{*}{Disease}	&	\multirow{2}{*}{Region}	&	\multirow{2}{*}{Naive}	&	\multirow{2}{*}{DLinear}	&	Random	&	\multirow{2}{*}{XGBoost}	&	\multirow{2}{*}{TSMixer}	&	\multirow{2}{*}{KAN}	&	\multirow{2}{*}{LSTM}	&	\multirow{2}{*}{DeepAR}	&	\multirow{2}{*}{TCN}	&	\multirow{2}{*}{NBeats}	&	\multirow{2}{*}{NHiTS}	&	Trans-	&	\multirow{2}{*}{TiDE}	&	\multirow{2}{*}{Chronos-2}	&	\multirow{2}{*}{TimesFM}	\\
    &	&	&	& Forest & & & & &	&	& 	&	&	former	&	&	&	\\  \midrule

    \multirow{20}{*}{Covid}	&	Australia	&	3.903	&	0.645	&	0.179	&	0.167	&	0.362	&	0.499	&	0.667	&	0.652	&	0.344	&	1.531	&	1.355	&	4.124	&	0.572	&	\underline{\textbf{0.115}}	&	\textit{0.127}	\\
	&	Belgium	&	2.370	&	0.046	&	0.036	&	0.040	&	0.388	&	0.084	&	0.060	&	0.064	&	0.054	&	0.105	&	0.031	&	2.955	&	0.042	&	\underline{\textbf{0.028}}	&	\textit{0.029}	\\
	&	Brazil	&	1.848	&	1.488	&	1.168	&	1.062	&	3.328	&	0.973	&	3.703	&	3.692	&	1.432	&	1.877	&	1.285	&	3.323	&	1.372	&	\underline{\textbf{0.888}}	&	\textit{0.906}	\\
	&	Canada 2	&	3.489	&	1.137	&	\textit{0.167}	&	0.258	&	0.829	&	0.345	&	0.385	&	0.385	&	0.821	&	9.595	&	2.878	&	0.863	&	1.879	&	\underline{\textbf{0.125}}	&	0.188	\\
	&	China	&	1.816	&	2.256	&	1.004	&	0.619	&	0.741	&	2.124	&	0.887	&	1.406	&	10.994	&	11.442	&	3.306	&	0.828	&	10.715	&	\underline{\textbf{0.336}}	&	\textit{0.439}	\\
	&	Colombia	&	6.205	&	10.619	&	\underline{\textbf{4.451}}	&	5.778	&	6.575	&	\textit{4.540}	&	7.026	&	7.014	&	140.355	&	77.501	&	42.152	&	6.636	&	44.974	&	4.748	&	4.744	\\
\bottomrule
        \end{tabular}
    \label{table_mase_medium}
\end{table}

\begin{table}[]
    \centering
    \caption{Medium-term forecasting performance of baseline and state-of-the-art models evaluated using the RMSE metric. The best results are indicated in \underline{\textbf{bold}}, while the second-best results are \textit{highlighted}. The average rank of each model for the specific forecasting task is reported at the end of the table.}
    \fontsize{5.49}{7}\selectfont
    \setlength{\tabcolsep}{0.5pt}
    \begin{tabular}{cccccccccccccccccccccc}
    \toprule
    \multirow{2}{*}{Disease}	&	\multirow{2}{*}{Region}	&	\multirow{2}{*}{Naive}	&	\multirow{2}{*}{DLinear}	&	Random	&	\multirow{2}{*}{XGBoost}	&	\multirow{2}{*}{TSMixer}	&	\multirow{2}{*}{KAN}	&	\multirow{2}{*}{LSTM}	&	\multirow{2}{*}{DeepAR}	&	\multirow{2}{*}{TCN}	&	\multirow{2}{*}{NBeats}	&	\multirow{2}{*}{NHiTS}	&	Trans-	&	\multirow{2}{*}{TiDE}	&	\multirow{2}{*}{Chronos-2}	&	\multirow{2}{*}{TimesFM}	\\
    &	&	&	& Forest & & & & &	&	& 	&	&	former	&	&	&	\\  \midrule
    \multirow{21}{*}{Covid}	&	Australia	&	162.044	&	21.193	&	7.427	&	9.740	&	11.475	&	6.830	&	15.271	&	14.616	&	9.650	&	6.550	&	7.581	&	169.359	&	7.201	&	\underline{\textbf{5.259}}	&	\textit{5.613}	\\
	&	Belgium	&	292.267	&	7.000	&	5.973	&	7.500	&	35.423	&	12.865	&	6.669	&	6.588	&	7.714	&	15.799	&	\underline{\textbf{4.398}}	&	353.591	&	6.164	&	4.988	&	\textit{4.939}	 \\
	&	Brazil	&	1444.375	&	1006.426	&	794.417	&	796.277	&	2451.869	&	750.762	&	2563.125	&	2559.977	&	1008.555	&	1005.368	&	855.777	&	2454.894	&	994.938	&	\textit{674.882}	&	\underline{\textbf{656.310}}	\\
	&	Canada	&	337.665	&	191.095	&	149.830	&	153.658	&	382.654	&	126.535	&	399.563	&	399.694	&	207.001	&	179.799	&	158.592	&	383.417	&	150.202	&	\textit{119.150}	&	\underline{\textbf{114.724}}	 \\
	&	Canada 2	&	1605.223	&	253.539	&	115.389	&	226.837	&	145.382	&	\textit{85.711}	&	156.868	&	156.463	&	107.726	&	665.377	&	421.385	&	145.858	&	234.185	&	\underline{\textbf{76.281}}	&	113.850	\\
	&	Chile	&	\underline{\textbf{4.670}}	&	15.413	&	6.781	&	7.346	&	5.665	&	22.810	&	5.864	&	5.882	&	8.161	&	17.715	&	14.312	&	5.645	&	19.488	&	\textit{4.871}	&	6.583	 \\
\bottomrule
            \end{tabular}
    \label{table_rmse_medium}
\end{table}

\begin{table}[]
    \centering
    \caption{Medium-term forecasting performance of baseline and state-of-the-art models evaluated using the MAE metric. The best results are indicated in \underline{\textbf{bold}}, while the second-best results are \textit{highlighted}. The average rank of each model for the specific forecasting task is reported at the end of the table.}
    \fontsize{5.49}{7}\selectfont
    \setlength{\tabcolsep}{0.5pt}
    \begin{tabular}{cccccccccccccccccccccc}
    \toprule
    \multirow{2}{*}{Disease}	&	\multirow{2}{*}{Region}	&	\multirow{2}{*}{Naive}	&	\multirow{2}{*}{DLinear}	&	Random	&	\multirow{2}{*}{XGBoost}	&	\multirow{2}{*}{TSMixer}	&	\multirow{2}{*}{KAN}	&	\multirow{2}{*}{LSTM}	&	\multirow{2}{*}{DeepAR}	&	\multirow{2}{*}{TCN}	&	\multirow{2}{*}{NBeats}	&	\multirow{2}{*}{NHiTS}	&	Trans-	&	\multirow{2}{*}{TiDE}	&	\multirow{2}{*}{Chronos-2}	&	\multirow{2}{*}{TimesFM}	\\
    &	&	&	& Forest & & & & &	&	& 	&	&	former	&	&	&	\\  \midrule
    \multirow{21}{*}{Covid}	&	Australia	&	161.947	&	18.470	&	6.305	&	7.123	&	10.022	&	5.573	&	13.764	&	13.520	&	7.924	&	5.378	&	6.120	&	169.263	&	5.821	&	\underline{\textbf{4.336}}	&	\textit{4.715}	\\
	&	Belgium	&	292.225	&	5.491	&	4.786	&	5.226	&	34.443	&	10.175	&	5.724	&	5.731	&	6.452	&	13.262	&	\underline{\textbf{3.438}}	&	353.558	&	5.017	&	\textit{3.778}	&	3.845	\\
	&	Brazil	&	1262.166	&	811.509	&	652.096	&	632.513	&	2245.789	&	600.513	&	2360.452	&	2357.127	&	827.638	&	815.846	&	704.103	&	2244.422	&	819.917	&	\textit{529.544}	&	\underline{\textbf{516.692}}	\\
	&	Canada	&	316.986	&	166.223	&	129.327	&	130.243	&	362.256	&	109.545	&	378.385	&	378.511	&	179.726	&	147.560	&	136.027	&	362.931	&	125.405	&	\textit{99.195}	&	\underline{\textbf{93.235}}	\\
	&	Canada 2	&	1604.170	&	231.445	&	105.980	&	177.846	&	137.347	&	\textit{74.305}	&	146.834	&	146.479	&	91.656	&	568.766	&	363.808	&	137.706	&	197.086	&	\underline{\textbf{64.170}}	&	98.274	\\
	&	Chile	&	\underline{\textbf{3.704}}	&	13.870	&	5.768	&	6.067	&	4.399	&	19.960	&	4.476	&	4.461	&	6.886	&	15.426	&	11.885	&	4.362	&	17.213	&	\textit{3.976}	&	5.385	\\
\bottomrule

        \end{tabular}
    \label{table_mae_medium}
\end{table}

\begin{table}[]
    \centering
    \caption{Medium-term forecasting performance of baseline and state-of-the-art models evaluated using the SMAPE metric. The best results are indicated in \underline{\textbf{bold}}, while the second-best results are \textit{highlighted}. The average rank of each model for the specific forecasting task is reported at the end of the table.}
    \fontsize{5.85}{7}\selectfont
    \setlength{\tabcolsep}{1pt}
    \begin{tabular}{cccccccccccccccccccccc}
    \toprule
    \multirow{2}{*}{Disease}	&	\multirow{2}{*}{Region}	&	\multirow{2}{*}{Naive}	&	\multirow{2}{*}{DLinear}	&	Random	&	\multirow{2}{*}{XGBoost}	&	\multirow{2}{*}{TSMixer}	&	\multirow{2}{*}{KAN}	&	\multirow{2}{*}{LSTM}	&	\multirow{2}{*}{DeepAR}	&	\multirow{2}{*}{TCN}	&	\multirow{2}{*}{NBeats}	&	\multirow{2}{*}{NHiTS}	&	Trans-	&	\multirow{2}{*}{TiDE}	&	\multirow{2}{*}{Chronos-2}	&	\multirow{2}{*}{TimesFM}	\\
    &	&	&	& Forest & & & & &	&	& 	&	&	former	&	&	&	\\  \midrule
    \multirow{21}{*}{Covid}	&	Australia	&	178.892	&	158.486	&	74.907	&	84.431	&	144.295	&	89.943	&	108.000	&	108.423	&	87.927	&	78.460	&	79.820	&	179.730	&	79.167	&	\textit{71.966}	&	\underline{\textbf{49.774}}	\\
	&	Belgium	&	195.323	&	163.449	&	142.498	&	155.986	&	200.000	&	149.195	&	139.715	&	\textit{133.596}	&	147.669	&	153.426	&	137.739	&	196.107	&	138.412	&	147.483	&	\underline{\textbf{126.434}}	 \\
	&	Brazil	&	57.557	&	50.157	&	36.668	&	32.099	&	149.157	&	28.881	&	191.858	&	190.043	&	45.540	&	60.018	&	42.916	&	147.180	&	49.241	&	\underline{\textbf{26.202}}	&	\textit{26.984}	\\
	&	Canada	&	147.553	&	115.061	&	\textit{77.097}	&	94.925	&	167.448	&	117.252	&	175.509	&	176.336	&	129.125	&	132.509	&	132.062	&	168.069	&	126.947	&	99.878	&	\underline{\textbf{65.584}}	 \\
	&	Canada 2	&	175.604	&	181.603	&	\underline{\textbf{49.887}}	&	110.474	&	131.114	&	92.275	&	169.675	&	167.729	&	106.300	&	163.503	&	140.588	&	130.942	&	138.498	&	90.409	&	\textit{54.004}	\\
	&	Chile	&	154.672	&	160.531	&	\underline{\textbf{133.169}}	&	\textit{147.334}	&	170.857	&	165.515	&	163.721	&	175.809	&	170.411	&	174.651	&	167.309	&	167.345	&	169.371	&	166.678	&	150.932	 \\
\bottomrule
            \end{tabular}
    \label{table_smape_medium}
\end{table}

\begin{table}[]
    \centering
    \caption{Long-term forecasting performance of baseline and state-of-the-art models evaluated using the MASE metric. The best results are indicated in \underline{\textbf{bold}}, while the second-best results are \textit{highlighted}. The average rank of each model for the specific forecasting task is reported at the end of the table.}
    \fontsize{5.9}{7}\selectfont
    \setlength{\tabcolsep}{0.7pt}
    \begin{tabular}{cccccccccccccccccccccc}
    \toprule
    \multirow{2}{*}{Disease}	&	\multirow{2}{*}{Location}	&	\multirow{2}{*}{Naive}	&	\multirow{2}{*}{DLinear}	&	Random	&	\multirow{2}{*}{XGBoost}	&	\multirow{2}{*}{TSMixer}	&	\multirow{2}{*}{KAN}	&	\multirow{2}{*}{LSTM}	&	\multirow{2}{*}{DeepAR}	&	\multirow{2}{*}{TCN}	&	\multirow{2}{*}{NBeats}	&	\multirow{2}{*}{NHiTS}	&	Trans-	&	\multirow{2}{*}{TiDE}	&	\multirow{2}{*}{Chronos-2}	&	\multirow{2}{*}{TimesFM}	\\
    &	&	&	& Forest & & & & &	&	& 	&	&	former	&	&	&	\\  \midrule
    \multirow{18}{*}{Covid}	&	Australia	&	3.843	&	1.502	&	0.198	&	0.180	&	0.442	&	0.852	&	0.619	&	0.813	&	0.432	&	3.675	&	1.297	&	6.403	&	0.557	&	\textit{0.175}	&	\underline{\textbf{0.174}}	\\
	&	Belgium	&	2.358	&	0.046	&	0.074	&	0.069	&	0.209	&	0.213	&	0.049	&	0.206	&	0.064	&	0.227	&	0.043	&	2.432	&	\textit{0.041}	&	\underline{\textbf{0.036}}	&	\underline{\textbf{0.036}}	\\
	&	Brazil	&	1.753	&	1.855	&	1.221	&	1.233	&	3.209	&	1.159	&	3.603	&	3.588	&	1.416	&	1.873	&	1.455	&	3.208	&	1.431	&	\textit{1.024}	&	\underline{\textbf{1.022}}	\\
	&	Canada 2	&	3.472	&	1.306	&	\underline{\textbf{0.255}}	&	0.317	&	0.875	&	0.476	&	0.394	&	0.398	&	1.045	&	3.955	&	1.468	&	0.861	&	1.977	&	0.338	&	\textit{0.308}	\\
	&	China	&	1.705	&	2.706	&	0.560	&	0.560	&	0.596	&	1.950	&	1.124	&	0.692	&	18.003	&	3.130	&	2.319	&	0.633	&	3.172	&	\underline{\textbf{0.438}}	&	\textit{0.448}	\\
	&	Colombia	&	8.699	&	12.946	&	\textit{7.406}	&	7.968	&	9.037	&	12.232	&	9.374	&	9.241	&	148.467	&	55.282	&	39.480	&	9.040	&	46.014	&	7.419	&	\underline{\textbf{7.356}}	\\
\bottomrule
            \end{tabular}
    \label{table_mase_long}
\end{table}

\begin{table}[]
    \centering
    \caption{Long-term forecasting performance of baseline and state-of-the-art models evaluated using the RMSE metric. The best results are indicated in \underline{\textbf{bold}}, while the second-best results are \textit{highlighted}. The average rank of each model for the specific forecasting task is reported at the end of the table.}
    \fontsize{5.44}{7}\selectfont
    \setlength{\tabcolsep}{0.5pt}
    \begin{tabular}{cccccccccccccccccccccc}
    \toprule
    \multirow{2}{*}{Disease}	&	\multirow{2}{*}{Region}	&	\multirow{2}{*}{Naive}	&	\multirow{2}{*}{DLinear}	&	Random	&	\multirow{2}{*}{XGBoost}	&	\multirow{2}{*}{TSMixer}	&	\multirow{2}{*}{KAN}	&	\multirow{2}{*}{LSTM}	&	\multirow{2}{*}{DeepAR}	&	\multirow{2}{*}{TCN}	&	\multirow{2}{*}{NBeats}	&	\multirow{2}{*}{NHiTS}	&	Trans-	&	\multirow{2}{*}{TiDE}	&	\multirow{2}{*}{Chronos-2}	&	\multirow{2}{*}{TimesFM}	\\
    &	&	&	& Forest & & & & &	&	& 	&	&	former	&	&	&	\\  \midrule
    \multirow{20}{*}{Covid}	&	Australia	&	161.785	&	11.712	&	\underline{\textbf{8.293}}	&	8.909	&	14.279	&	17.210	&	15.264	&	18.567	&	11.980	&	23.794	&	9.103	&	255.498	&	\textit{8.302}	&	8.580	&	9.050	\\ 
	&	Belgium	&	294.023	&	7.464	&	14.714	&	14.127	&	17.923	&	31.756	&	8.580	&	23.275	&	9.707	&	39.505	&	\underline{\textbf{5.899}}	&	297.278	&	\textit{6.201}	&	7.073	&	7.123	\\
	&	Brazil	&	1425.234	&	1098.312	&	900.877	&	1108.051	&	2383.400	&	837.855	&	2496.986	&	2493.070	&	1079.979	&	1098.713	&	970.464	&	2388.117	&	977.751	&	\textit{766.134}	&	\underline{\textbf{759.863}}	\\
	&	Canada	&	311.056	&	260.149	&	193.669	&	190.364	&	348.285	&	\underline{\textbf{162.857}}	&	360.995	&	360.690	&	213.884	&	216.034	&	\textit{175.884}	&	349.054	&	194.941	&	185.533	&	178.036	\\ 
	&	Canada 2	&	1632.683	&	291.587	&	209.978	&	267.736	&	\textit{145.833}	&	147.091	&	159.542	&	158.834	&	\underline{\textbf{133.44}}	&	288.952	&	180.798	&	146.470	&	288.589	&	233.947	&	259.464	\\
	&	Chile	&	7.109	&	6.691	&	6.030	&	6.178	&	7.445	&	32.098	&	7.711	&	7.604	&	8.827	&	61.898	&	33.430	&	7.545	&	32.474	&	\underline{\textbf{4.114}}	&	\textit{4.551}	\\
\bottomrule
    \end{tabular}
    \label{table_rmse_long}
\end{table}

\begin{table}[]
    \centering
    \caption{Long-term forecasting performance of baseline and state-of-the-art models evaluated using the MAE metric. The best results are indicated in \underline{\textbf{bold}}, while the second-best results are \textit{highlighted}. The average rank of each model for the specific forecasting task is reported at the end of the table.}
    \fontsize{5.49}{7}\selectfont
    \setlength{\tabcolsep}{0.5pt}
    \begin{tabular}{cccccccccccccccccccccc}
    \toprule
    \multirow{2}{*}{Disease}	&	\multirow{2}{*}{Region}	&	\multirow{2}{*}{Naive}	&	\multirow{2}{*}{DLinear}	&	Random	&	\multirow{2}{*}{XGBoost}	&	\multirow{2}{*}{TSMixer}	&	\multirow{2}{*}{KAN}	&	\multirow{2}{*}{LSTM}	&	\multirow{2}{*}{DeepAR}	&	\multirow{2}{*}{TCN}	&	\multirow{2}{*}{NBeats}	&	\multirow{2}{*}{NHiTS}	&	Trans-	&	\multirow{2}{*}{TiDE}	&	\multirow{2}{*}{Chronos-2}	&	\multirow{2}{*}{TimesFM}	\\
    &	&	&	& Forest & & & & &	&	& 	&	&	former	&	&	&	\\  \midrule
    \multirow{20}{*}{Covid}	&	Australia	&	161.565	&	9.755	&	6.764	&	\textit{6.558}	&	12.547	&	14.309	&	13.671	&	17.373	&	10.092	&	19.373	&	6.869	&	255.358	&	6.789	&	\underline{\textbf{6.384}}	&	6.581	\\
	&	Belgium	&	293.961	&	5.975	&	10.843	&	10.623	&	16.195	&	24.192	&	6.391	&	18.841	&	7.867	&	29.527	&	\underline{\textbf{4.726}}	&	297.216	&	4.982	&	\textit{4.783}	&	4.820	\\
	&	Brazil	&	1209.790	&	879.730	&	715.257	&	745.527	&	2148.292	&	678.635	&	2266.177	&	2261.962	&	866.836	&	850.653	&	735.926	&	2148.058	&	781.227	&	\underline{\textbf{582.216}}	&	\textit{583.966}	\\ 
	&	Canada	&	286.335	&	212.528	&	166.916	&	161.327	&	324.021	&	\underline{\textbf{132.743}}	&	336.128	&	335.860	&	181.536	&	171.141	&	\textit{141.467}	&	324.777	&	158.230	&	159.365	&	149.583	\\
	&	Canada 2	&	1631.464	&	253.297	&	191.388	&	214.250	&	136.782	&	\textit{121.927}	&	149.462	&	148.778	&	\underline{\textbf{111.508}}	&	259.824	&	161.942	&	137.701	&	252.218	&	201.789	&	235.180	\\
	&	Chile	&	6.147	&	6.065	&	5.288	&	5.393	&	6.485	&	21.831	&	6.683	&	6.583	&	7.578	&	41.235	&	22.515	&	6.551	&	21.808	&	\underline{\textbf{3.258}}	&	\textit{3.930}	\\
\bottomrule
        \end{tabular}
    \label{table_mae_long}
\end{table}

\begin{table}[]
    \centering
    \caption{Long-term forecasting performance of baseline and state-of-the-art models evaluated using the SMAPE metric. The best results are indicated in \underline{\textbf{bold}}, while the second-best results are \textit{highlighted}. The average rank of each model for the specific forecasting task is reported at the end of the table.}
    \fontsize{6}{7}\selectfont
    \setlength{\tabcolsep}{0.7pt}
    \begin{tabular}{cccccccccccccccccccccc}
    \toprule
        \multirow{2}{*}{Disease}	&	\multirow{2}{*}{Region}	&	\multirow{2}{*}{Naive}	&	\multirow{2}{*}{DLinear}	&	Random	&	\multirow{2}{*}{XGBoost}	&	\multirow{2}{*}{TSMixer}	&	\multirow{2}{*}{KAN}	&	\multirow{2}{*}{LSTM}	&	\multirow{2}{*}{DeepAR}	&	\multirow{2}{*}{TCN}	&	\multirow{2}{*}{NBeats}	&	\multirow{2}{*}{NHiTS}	&	Trans-	&	\multirow{2}{*}{TiDE}	&	\multirow{2}{*}{Chronos-2}	&	\multirow{2}{*}{TimesFM}	\\
    &	&	&	& Forest & & & & &	&	& 	&	&	former	&	&	&	\\  \midrule
   \multirow{20}{*}{Covid}	&	Australia	&	173.686	&	104.206	&	\textit{74.472}	&	77.539	&	147.249	&	102.593	&	102.405	&	110.523	&	87.462	&	102.695	&	74.771	&	182.700	&	76.062	&	76.127	&	\underline{\textbf{51.662}}	\\
	&	Belgium	&	193.437	&	166.271	&	148.414	&	152.388	&	185.476	&	152.390	&	185.552	&	144.164	&	\textit{131.277}	&	153.927	&	\underline{\textbf{129.019}}	&	193.510	&	131.591	&	172.901	&	154.816	\\
	&	Brazil	&	55.770	&	54.586	&	39.342	&	37.153	&	147.591	&	34.663	&	193.235	&	190.823	&	44.578	&	58.096	&	47.894	&	145.840	&	44.924	&	\textit{32.482}	&	\underline{\textbf{31.872}}	\\
	&	Canada	&	151.303	&	130.239	&	\underline{\textbf{97.336}}	&	113.086	&	170.749	&	125.252	&	175.878	&	175.810	&	147.930	&	141.211	&	137.135	&	170.529	&	132.517	&	136.923	&	\textit{99.637}	\\
	&	Canada 2	&	174.244	&	170.432	&	\underline{\textbf{78.501}}	&	\textit{99.462}	&	126.036	&	112.501	&	172.682	&	169.253	&	105.448	&	133.634	&	113.747	&	127.660	&	129.035	&	107.517	&	\underline{\textbf{57.664}}	\\
	&	Chile	&	164.234	&	153.497	&	\textit{150.101}	&	\underline{\textbf{148.907}}	&	177.380	&	160.246	&	176.049	&	179.505	&	176.493	&	175.847	&	163.476	&	175.942	&	172.942	&	172.882	&	159.124	\\
\bottomrule
    \end{tabular}
    \label{table_smape_long}
\end{table}

\subsection{Empirical results}
Figure \ref{Fig_Boxplot_SMAPE_MAE} summarizes the performance of the forecasting models across different horizons using the SMAPE and MAE metrics. The box plots highlight that the foundation models outperform the competitive forecasting approaches across diverse epidemic datasets, aligning with patterns observed for the MASE and RMSE metrics. Additionally, detailed empirical results for each model and dataset, evaluated across all performance metrics, are reported in Tables \ref{table_mase_short}–\ref{table_smape_long}.

\begin{figure}[t]
    \centering
    \begin{subfigure}{\linewidth}
        \centering
        \includegraphics[width=\linewidth]{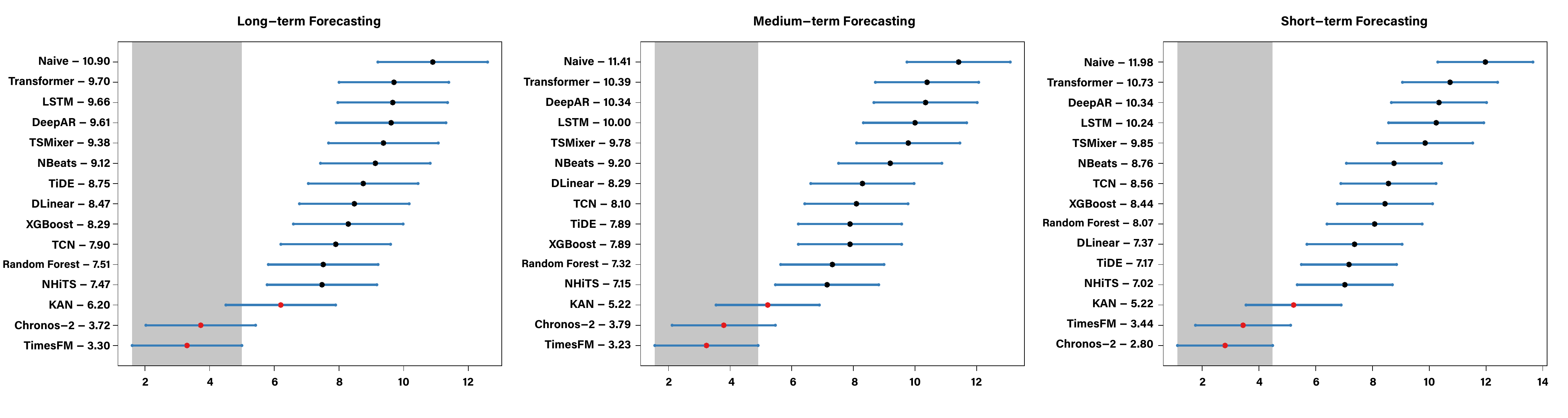}
    (a) MCB test based on RMSE metric
    \end{subfigure}

    \begin{subfigure}{\linewidth}
        \centering
        \includegraphics[width=\linewidth]{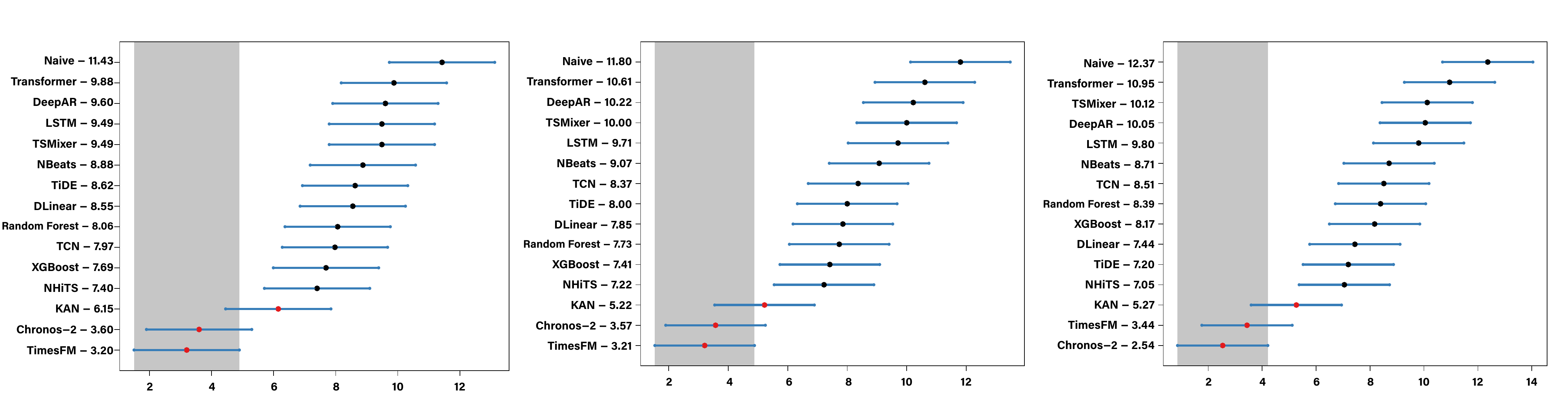}
    (b) MCB test based on MAE metric
    \end{subfigure}

    \begin{subfigure}{\linewidth}
        \centering
        \includegraphics[width=\linewidth]{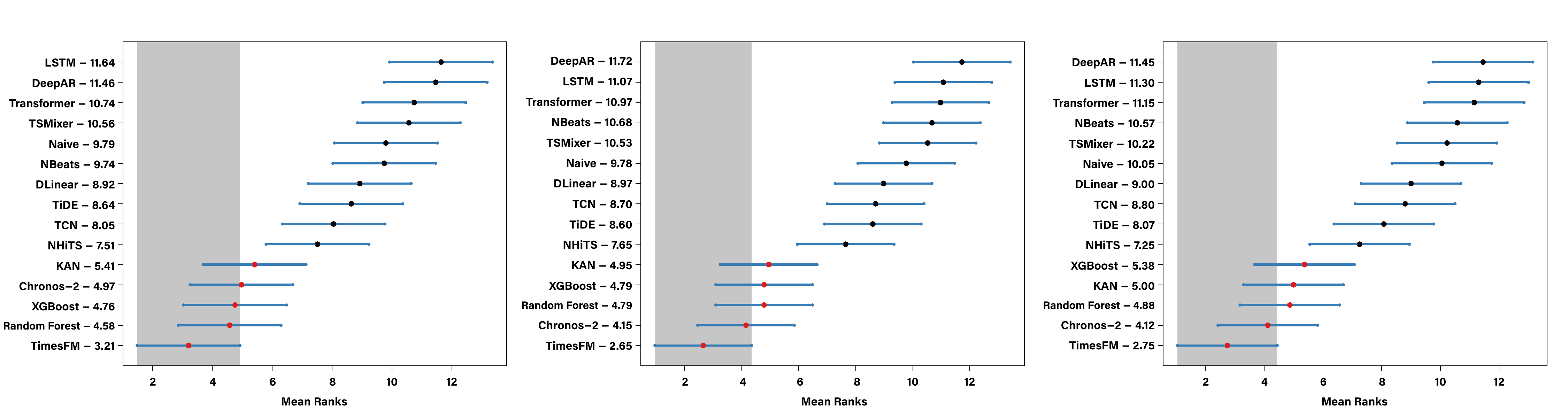}
    (c) MCB test based on SMAPE metric
    \end{subfigure}

    \caption{Multiple comparisons with the best (MCB) test based on the (a) RMSE, (b) MAE, and (c) SMAPE metrics across long (left), medium (middle), and short-term (right) forecasting horizons. In the plot, `TimesFM - 3.30' indicates the average rank of the TimesFM framework for long-term forecasting is 3.30 based on the RMSE metric; a similar interpretation holds for other models, metrics and horizons.}   
    \label{MCB_rest_Plot}
\end{figure}

To analyze individual model performance across different forecasting settings, we examine their capabilities along multiple dimensions. First, to assess how performance varies with the size of the epidemic time series, we group the datasets into three categories: small (fewer than 100 observations), moderate (100-500 observations), and relatively large (more than 500 observations). The datasets cover diverse epidemiological regimes, including airborne infections (COVID-19, measles, and tuberculosis), vector-borne diseases (dengue, chikungunya, Zika), contact-based diseases (chickenpox), and droplet-borne diseases (influenza), with temporal resolutions ranging from daily to weekly and monthly reporting. A key characteristic of the epidemic data is the presence of zero inflation, particularly in several COVID-19 datasets. This is most pronounced in countries such as China, Chile, Canada, Australia, Colombia, and the US, with relatively lower sparsity observed in datasets from the UK, Belgium, Ireland, Italy, Japan, the EU, Czech, and Spain. In our analysis, we investigate the joint impact of dataset size and zero inflation on model performance. We further interpret these results in light of the global time series characteristics which vary across disease types and directly influence model suitability. Across forecasting horizons, the focus remains on identifying consistent performance patterns rather than isolated successes.

The short-term forecasting performance, presented in Tables \ref{table_mase_short}-\ref{table_smape_short}, shows clear patterns across dataset sizes and disease types. For small sample-sized datasets, foundation models account for the majority of top-performing results, highlighting their robustness in data-scarce settings. Among them, Chronos-2 more frequently achieves the lowest errors, although its advantage over TimesFM is typically marginal. Notable deviations arise in highly zero-inflated settings. For instance, in the COVID-19 dataset from Chile, Random Forest and DeepAR outperform foundation models, suggesting that under severe sparsity and limited data, ensemble and recurrent approaches may better capture the underlying dynamics. A similar but less pronounced pattern is observed for chikungunya cases of Brazil, where TSMixer and TiDE slightly outperform foundation models, though the performance gap remains small. Notably, Chronos-2 maintains strong performance even in extremely small datasets; for example, in the tuberculosis dataset of China with only 60 observations, it remains highly competitive, due to its large-scale pretraining, which enables it to generalize effectively despite limited data availability. For moderately sized datasets, foundation models continue to perform strongly, particularly for COVID-19. Chronos-2 and TimesFM alternate as the top-performing models across countries, including zero-inflated cases such as China and Canada, where both maintain stable performance. While metric-specific differences exist, they are generally modest; for example, KAN achieves the best scale-dependent performance for Colombia’s COVID-19 dataset but remains close to both foundation models. A similar trend is observed for dengue, where foundation models lead in several cases but do not dominate uniformly, with models such as DLinear and KAN outperforming them for the Philippines data. This behavior aligns with the combination of high spikiness, strong trend components, and moderate long memory observed in vector-borne diseases, where abrupt outbreaks and persistent growth patterns reduce the effectiveness of global sequence representations and favor models that adapt to localized dynamics. More consistent rankings appear in certain scenarios, such as for chikungunya cases of Colombia, where TimesFM consistently performs best, and Zika in Mexico, where Chronos-2 leads across most metrics. In contrast, for the tuberculosis dataset of Japan, a distinct pattern is observed, with the deep learning-based NHiTS model outperforming other competitive architectures, likely due to its hierarchical design, which is well suited to capturing long memory effects. For relatively large datasets, performance again concentrates among foundation models. Chronos-2 and TimesFM jointly dominate COVID-19 forecasting across different countries, including highly zero-inflated datasets such as Australia and the US, although neither consistently outperforms the other across all metrics. Some localized deviations persist, for instance, TiDE and Random Forest perform better for Japan, DeepAR and KAN for the EU dataset, and NHiTS provides competitive forecasts for COVID-19 incidence in Czech. For other diseases, clearer model-specific patterns emerge, with TiDE performing best for chickenpox, Chronos-2 leading for dengue cases of Colombia, TimesFM dominating influenza forecasting in the US, and Chronos-2 generating the most reliable forecasts for measles. In particular, for contact-based diseases, the combination of high entropy, strong seasonality, and pronounced curvature alongside moderate trend components creates a structure where periodic behavior coexists with substantial variability and smoother transitions. In this setting, models such as TiDE or KAN can outperform foundation models by better capturing smooth seasonal dynamics while remaining sensitive to local deviations, as observed for chickenpox in Hungary. Across all dataset sizes, the Naive model and several deep learning architectures, including Transformer, consistently underperform, particularly in zero-inflated COVID-19 datasets. This trend persists in moderate and relatively large datasets, such as those from Australia, China, and Canada, indicating that sparsity remains a key challenge for these models. However, in small and highly sparse datasets, such as COVID-19 in Chile, conventional machine learning models can outperform foundation models, suggesting that the combination of limited data and zero inflation poses a critical challenge in short-term forecasting tasks.

The medium-term forecasting results, presented in Tables \ref{table_mase_medium}-\ref{table_smape_medium}, reveal distinct patterns across dataset sizes and disease types. For small sample-sized datasets, TimesFM emerges as the most consistent framework for chikungunya, dengue, Zika, and tuberculosis incidence cases. Notable deviations arise in COVID-19 datasets from Spain and Chile, where models such as TSMixer and Random Forest outperform foundation models, suggesting that irregular and highly variable epidemic patterns can favor alternative approaches, consistent with observations in short-term forecasting. For moderately sized datasets, performance becomes more heterogeneous and increasingly dependent on the underlying disease dynamics. In COVID-19 cases, Chronos-2 and TimesFM continue to dominate, alternating as the top-performing models across datasets and metrics, including zero-inflated settings. A localized deviation is observed for Colombia, where Random Forest and KAN provide competitive forecasts. In contrast, vector-borne diseases exhibit greater variability, with dengue showing no consistent ranking pattern. However, more stable behavior emerges for chikungunya in Colombia and Zika in Mexico, where foundation models maintain strong performance. Tuberculosis in Japan again presents a distinct case, with NHiTS outperforming other competitive approaches. For relatively large sample-sized datasets, performance becomes more structured. Foundation models, particularly TimesFM, outperform across most COVID-19 datasets, with Chronos-2 remaining highly competitive, especially in zero-inflated cases. Tree-based models such as Random Forest and XGBoost yield occasional improvements but do not generalize consistently across datasets and metrics. Similar trends extend to other diseases, where foundation models dominate the influenza dataset and remain competitive for measles cases in the US. Persistent deviations include chickenpox in Hungary, where KAN consistently outperforms other approaches. In dengue data from Colombia, Chronos-2 performs best under scale-free evaluation, while KAN outperforms under scale-dependent metrics.

Results for the long-term forecasting horizon, presented in Tables \ref{table_mase_long}–\ref{table_smape_long}, reveal clear patterns across dataset sizes and disease types. For small datasets, foundation models consistently produce more accurate and reliable forecasts, with Chronos-2 and TimesFM accounting for most top-performing results across diseases. Differences between the two frameworks remain dataset and metric-specific, with no consistent ranking. In the zero-inflated COVID-19 dataset from Chile, for instance, Chronos-2 achieves the strongest performance on scale-dependent metrics, contrasting with the short-term horizon where ensemble models performed better. This suggests that the advantage of foundation models under zero inflation becomes more pronounced as the forecasting horizon increases. Deviations are primarily observed in low-frequency settings with limited samples, such as tuberculosis cases of China, where ensemble models like Random Forest and XGBoost, along with linear architectures such as DLinear, outperform more complex approaches. This highlights the sensitivity of long-term forecasting to limited data and coarse temporal resolution. For moderately sized datasets, the strong performance of foundation models persists, particularly for COVID-19. The metric-specific deviations are observed; for example, KAN and TCN provide occasional improvements for the Canada dataset, while LSTM yields competitive results for Switzerland. In contrast, vector-borne diseases exhibit greater variability. For chikungunya cases of Colombia, model performance depends largely on the evaluation metric, with TimesFM leading on scale-free measures and Chronos-2 performing better for scale-dependent ones. Dengue datasets show heterogeneous outcomes, with foundation models and the DeepAR framework achieving top performance depending on the dataset and metric. More stable rankings are observed for Zika incidences in Mexico, where TimesFM consistently performs best. Tuberculosis in Japan again presents mixed behavior, with NHiTS and KAN outperforming other approaches across different metrics. For relatively large datasets, a clearer hierarchy emerges with TimesFM achieving the best overall performance across most COVID-19 datasets. Alternative forecasting approaches such as TiDE, KAN, and Random Forest provide dataset-specific improvements, but these gains are not consistent across settings. For other diseases, foundation models generally outperform alternative forecasting approaches, although persistent exceptions remain, such as chickenpox, where KAN dominates for most metrics.

Across all forecasting horizons, several consistent patterns emerge. First, Chronos-2 and TimesFM form the dominant pair for different datasets, with their zero-inflation robustness becoming more pronounced as the forecasting horizon increases. This is particularly evident in airborne and droplet-borne diseases, where stronger seasonality, stable trend evolution, and moderate long memory provide sufficient global structure for foundation models to effectively capture persistent temporal dependencies, leading to stable performance across datasets. 
Second, the advantage of foundation models is modulated by the interaction between zero inflation and sample size. While foundation models handle sparsity effectively in larger datasets, their advantage can diminish in small, highly zero-inflated series, where ensemble-based methods occasionally perform better. This effect is most evident in short-term forecasting. 
Third, disease-specific structure introduces additional variability in ranking stability. Vector-borne diseases such as dengue and chikungunya show the least stable behavior, driven by sharp peaks and irregular outbreak patterns that disrupt temporal continuity, resulting in inconsistent rankings where foundation models and frameworks such as KAN and Random Forest both alternate in dominance due to their complementary strengths in global representation and abrupt local adaptation. In contrast, airborne diseases such as COVID-19, measles, and tuberculosis exhibit more consistent dynamics with stable trends and long memory, where foundation models generally perform best by leveraging persistent temporal structure. Droplet-borne influenza shows smoother but still evolving patterns shaped by seasonality and curvature, again favoring foundation models for capturing periodic structure and medium-range dependencies. Contact-based chickenpox combines high entropy, strong seasonality, and pronounced curvature, where models such as TiDE and KAN occasionally outperform foundation models due to stronger local adaptation. Finally, certain model classes consistently underperform across different forecasting settings. The Naive model and several deep learning architectures, including Transformers and RNN-based frameworks, may achieve isolated gains but fail to generalize across datasets and forecasting horizons, particularly under significant zero inflation. Overall, forecasting performance is primarily driven by the interaction between dataset size and sparsity rather than any single factor. Foundation models provide the most stable performance across these dimensions, while alternative models contribute targeted improvements in specific settings.

Furthermore, Figure \ref{MCB_rest_Plot} presents the statistical robustness of model performance using the MCB test across all forecasting models based on RMSE, MAE, and SMAPE metrics. Each panel reports the mean rank of competing models along with the corresponding critical distance (CD), where lower ranks indicate better predictive performance. The shaded region identifies models that are statistically comparable with the best-performing approach. Results based on RMSE and MAE indicate that foundation models, particularly TimesFM, achieve the lowest average ranks for medium and long-term forecasting horizons, while Chronos-2 performs best in the short-term forecasting. In addition, their CDs largely overlap with the reference value of the test, highlighting their statistical competitiveness. This behaviour is primarily attributed to large-scale pretraining, which enables strong generalization even under limited task-specific data. Among conventional approaches, KAN demonstrates relatively strong performance, outperforming several machine learning and deep learning models; although it attains slightly higher (worse) mean ranks than foundation models, the overlapping CDs suggest that these differences are not statistically significant. The MCB results based on SMAPE reveal a slightly different ranking structure. While foundation models consistently outperform conventional approaches across all horizons, tree-based methods such as XGBoost and Random Forest, along with KAN, remain statistically competitive with them in several cases. In contrast, the Naive model exhibits consistently higher (worse) ranks due to its persistence-based forecasting strategy, which simply repeats the last observed value and fails to capture epidemic dynamics. Deep learning models show moderate but less stable performance across horizons. Transformer and RNN-based architectures, including LSTM and DeepAR, are particularly sensitive to dataset size, as they require substantial data to effectively learn temporal dependencies. The limited sample size of several epidemic datasets, therefore, poses a significant challenge for these models. In comparison, architectures such as NHiTS and TiDE demonstrate relatively more stable performance, although they exhibit significant performance differences from foundation models in these forecasting tasks. Notably, the relative ordering of models remains largely consistent across forecasting horizons, indicating robustness in comparative performance. Overall, the MCB analysis confirms the consistent superiority of foundation models, while also highlighting that KAN and tree-based ensembles can provide competitive alternatives without statistically significant gaps in specific epidemic forecasting scenarios.

\end{document}